\documentclass{article}

\PassOptionsToPackage{numbers}{natbib}
\usepackage[final]{neurips_data_2024}

\usepackage[utf8]{inputenc} % allow utf-8 input
\usepackage[T1]{fontenc}    % use 8-bit T1 fonts
\usepackage{hyperref}       % hyperlinks
\usepackage{url}            % simple URL typesetting
\usepackage{booktabs}       % professional-quality tables
\usepackage{amsfonts}       % blackboard math symbols
\usepackage{nicefrac}       % compact symbols for 1/2, etc.
\usepackage{microtype}      % microtypography
\usepackage{multirow}
\usepackage[table,xcdraw]{xcolor}
\usepackage{siunitx} % For handling numbers
\usepackage{graphicx}
\usepackage{caption}
\usepackage{subcaption}
\usepackage{amsmath}
\usepackage{algorithm}
\usepackage{algpseudocode}
\usepackage{tabularx}
\usepackage{tcolorbox}
\usepackage{adjustbox}
\usepackage{natbib}

\usepackage[normalem]{ulem}
\useunder{\uline}{\ul}{}

\definecolor{softviolet}{rgb}{0.7, 0.4, 0.8}
\definecolor{softblue}{rgb}{0.4, 0.6, 0.8}
\definecolor{forestgreen}{rgb}{0.13, 0.55, 0.13}
\definecolor{forestred}{rgb}{0.55, 0.13, 0.13}
\definecolor{lightgreen}{HTML}{C4F4C8}
\definecolor{lightblue}{HTML}{C4F4E0}
\definecolor{lightgrey}{HTML}{A4A4A4}
\definecolor{ForestGreen}{HTML}{228B22}

\usepackage{pifont}% http://ctan.org/pkg/pifont
\newcommand{\cmark}{\ding{51}}%
\newcommand{\xmark}{\ding{55}}%

\newcommand{\maintitleblock}{
    \title{CoMix: A Comprehensive Benchmark for Multi-Task Comic Understanding}
    
    \author{%
    Emanuele Vivoli$^{1,2}$ \quad Marco Bertini$^{2}$ \quad Dimosthenis Karatzas$^1$\\
    $^1$Computer Vision Center, UAB, Spain \quad $^2$MICC, University of Florence, Italy\\
    \texttt{\{evivoli,dimos\}@cvc.uab.cat}\\
    \texttt{\{emanuele.vivoli, marco.bertini\}@unifi.it}
    }

    \maketitle
}

\newcommand{\suptitleblock}{
    \vspace{0.5em}
    \noindent\hrule height 3pt
    \begin{center}
        {\LARGE \bfseries CoMix: A Comprehensive Benchmark for Multi-Task Comic Understanding \par}
    \end{center}
    \vspace{0.5em}
    \noindent\hrule height 1pt
    \vspace{0.5em}
    \begin{center}
        {\large Supplementary Material \par}
    \end{center}
    \vspace{1.5em}
}

\begin{document}

\maintitleblock

\begin{figure}[h!]
    \vspace{-1cm}
    \centering
    \includegraphics[height=6cm]{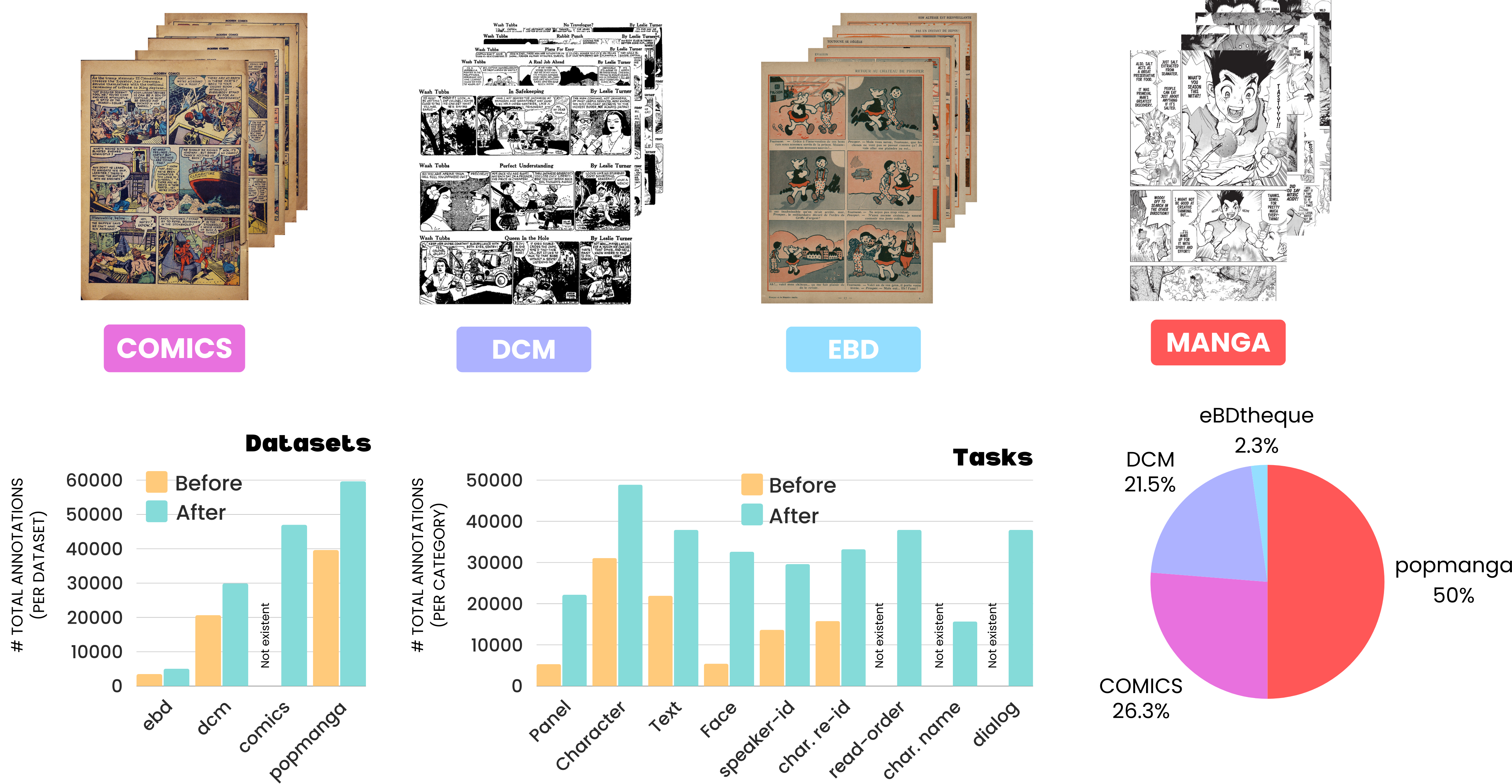}
    % \caption{Composition of the \textit{CoMix} benchmark. \textbf{Top} shows a qualitative representation of the datasets included in \textit{CoMix}. The bar charts illustrate the differences between existing annotations and the ones extended in \textit{CoMix}. \textbf{Left} shows increased number of annotations per dataset, while \textbf{right} per task.}
    \caption{Composition of the \textit{CoMix} benchmark. The \textbf{top} part of the figure provides a qualitative representation of the datasets included in \textit{CoMix}. The accompanying bar charts depict the differences between the original annotations and those extended in \textit{CoMix}. The \textbf{left} chart shows the increased number of annotations per dataset, whereas the \textbf{right} chart details the increase per task.}

    \label{fig:enter-label}
\end{figure}

\begin{abstract}
The comic domain is rapidly advancing with the development of single-page analysis and synthesis models. However, evaluation metrics and datasets lag behind, often limited to small-scale or single-style test sets. We introduce a novel benchmark, \textit{CoMix}, designed to evaluate the multi-task capabilities of models in comic analysis. Unlike existing benchmarks that focus on isolated tasks such as object detection or text recognition, \textit{CoMix} addresses a broader range of tasks including object detection, speaker identification, character re-identification, reading order, and multi-modal reasoning tasks like character naming and dialogue generation. Our benchmark comprises three existing datasets with expanded annotations to support multi-task evaluation. To mitigate the over-representation of manga-style data, we have incorporated a new dataset of carefully selected American comic-style books, thereby enriching the diversity of comic styles. \textit{CoMix} is designed to assess pre-trained models in zero-shot and limited fine-tuning settings, probing their transfer capabilities across different comic styles and tasks. The validation split of the benchmark is publicly available for research purposes, and an evaluation server for the held-out test split is also provided. Comparative results between human performance and state-of-the-art models reveal a significant performance gap, highlighting substantial opportunities for advancements in comic understanding. The dataset, baseline models, and code are accessible at \href{https://github.com/emanuelevivoli/CoMix-dataset}{the repository link}. This initiative sets a new standard for comprehensive comic analysis, providing the community with a common benchmark for evaluation on a large and varied set.
\end{abstract}

\section{Introduction}

Comics, a distinct medium that seamlessly blends textual and visual components, has emerged as a globally celebrated form of cultural expression. While the accessibility of comics allows even young readers to comprehend and appreciate them with ease, the intricacy of comic page layouts poses significant challenges for computational understanding. The classical elements of comics, including panels, speech balloons, characters, text, and onomatopoeia, are heavily shaped by the creator's imaginative vision and artistic flair, rendering the undertaking of comic image analysis a complex and multifaceted endeavor.

In recent works\citep{Vivoli2024OneMP}, authors approached comics research to more complex tasks moving from \textit{classification}\citep{Vivoli2024MultimodalTF}, \textit{detection}\citep{Vivoli2024ComicsDF} and \textit{captioning}\citep{Vivoli2024ComiCapAV} to \textit{diarization}\citep{li_manga109dialog_2023,sachdeva_manga_2024}, where low-level tasks - such as detecting objects, defining reading order and speaker identification and character re-identification - are used as a medium to generate an ordered transcription of who said what on single page. Despite the innovative application of this complex bottom-up approach to comics understanding tasks, there is no benchmark on character naming or diarization, nor metrics to assess their correctness \citep{Vivoli2024OneMP}.

A number of available datasets support some of these tasks. For example, eBDtheque \citep{guerin_ebdtheque_2013} and DCM772 \citep{nguyen_digital_2018} provide detection annotations for panels, characters, text lines, and some occurrences of character's faces. Another recent dataset called PopManga \citep{sachdeva_manga_2024} offers annotations to character and text detection as well as speaker identification and re-identification. Lastly, the well-known Manga109 dataset \citep{fujimoto_manga109_2016} has seen various iterations that enriched the annotations landscape from just object detection and character re-identification to the late onomatopoeia detection and recognition \citep{baek_coo_2022} and speaker identification \citep{li_manga109dialog_2023}. However, the most comprehensive dataset, Manga109, comprehends only manga-style Japanese comics, limiting a proper evaluation of generalization capability across different styles and lacking a diarization ground truth.

In this work, we propose the \textit{CoMix} - a benchmark formed of purposefully selected and annotated comic books that aims to comprehensively assess the capability of single and multimodal models across different vision tasks (detection, re-identification, OCR), multimodal tasks (speaker identification, character naming, reading order, dialog generation), spanning different types of comics style (American, Manga and small percentage of French) and modalities (allowing for single page and multi-page). Our benchmark draws inspiration from datasets like Manga109 \citep{fujimoto_manga109_2016}, and recent work like Magi \citep{sachdeva_manga_2024}, closing the gap between more capable models and datasets/metrics unavailability.

\begin{figure}
    \centering
    \includegraphics[height=8cm]{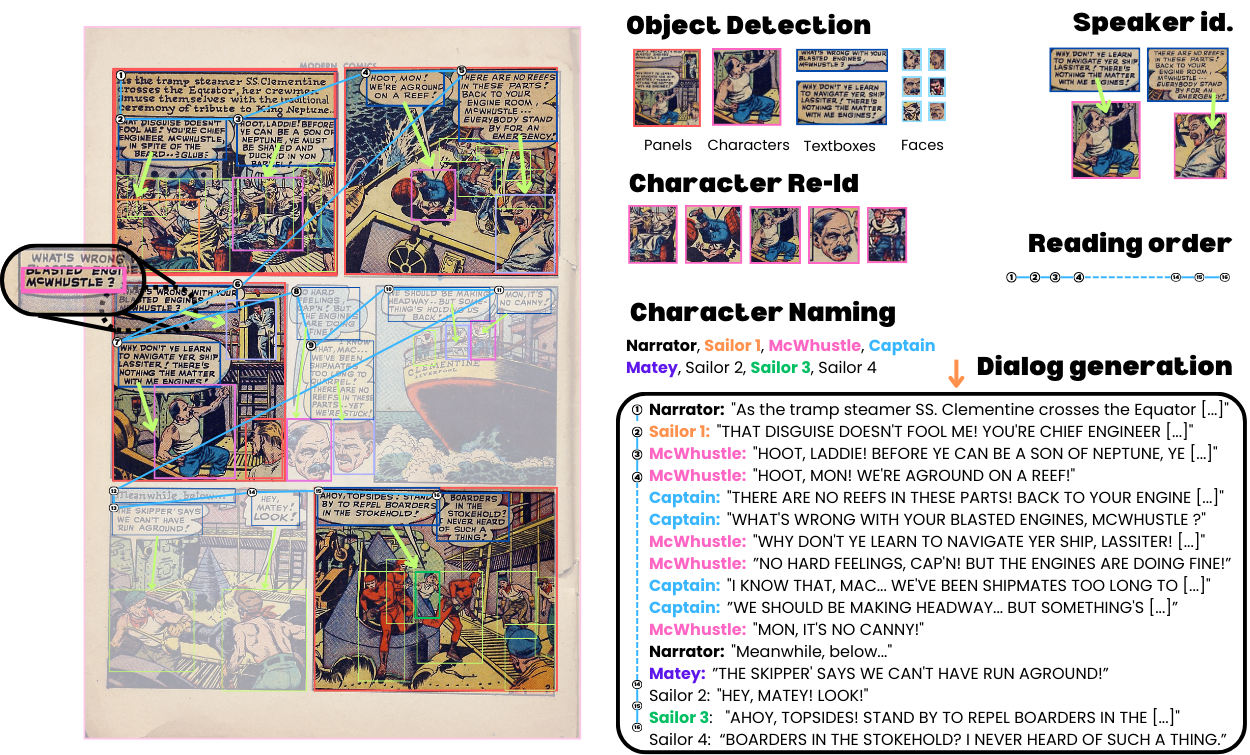}
    \caption{The \textit{CoMix} benchmark contains 4 computational tasks (object detection, speaker identification, character re-identification, panel-text sorting) and 2 multi-modal reasoning tasks (character naming and dialog generation) which require models to detect objects and their relation, as well as reading text. The figure shows the annotations added for each comic page, and on the left is depicted an example annotation of multi-modal reasoning task \textit{dialog generation}.}
    \label{fig:first-figure}
    \vspace{-5mm}
\end{figure}

\textit{CoMix} contains 3.8k images, gathered from 100 books, densely annotated with 130K objects across the four classes considered, 30k text-characters links, with 33k characters clusters with 16k total names identified. An overview of single-page annotations is provided in Figure \ref{fig:first-figure} while statistics about datasets and annotations are provided in Table \ref{tab:dataset-stats}.
We open-source the images and annotations of the validation splits. An evaluation server and images from the held-out test split are made available. Since currently there is no model that can tackle all the evaluation tasks in our benchmark, %(\citet{sachdeva_manga_2024}'s model does not identify names, and only tackles manga-style reading order), 
we provide baseline results for per-task models: object detection, speaker identification, character re-identification, reading order, character naming, and dialog generation.

The contributions of this work are as follows:
\begin{itemize}
    \item We address the lack of metrics and benchmark datasets for comics understanding, proposing high-level tasks metrics: Hybrid Dialog Score for character naming and dialog generation;
    \item We introduce \textit{CoMix}, a diverse manga- and comics-style benchmark of carefully selected books with computational and reasoning-dense annotations;
    \item We provide baseline results for each task, identifying substantial performance gaps with human baselines.
\end{itemize}

In the next section (section \ref{sec:related-work}), we discuss related work in more detail, highlighting what sets the \textit{CoMix} apart in the landscape of comics analysis datasets and benchmarks. In sections \ref{sec:books} and \ref{sec:anns}, we describe the books and annotations in the \textit{CoMix}, with details about the diversity of comics and artistic styles. In section \ref{sec:baseline-results}, we introduce the tasks enabled by these annotations, together with evaluation metrics and baselines, including a human baseline. We conclude with a summary and directions for future work in section \ref{sec:conclusion}.

\section{Related Work}
\label{sec:related-work}
\iffalse
\begin{itemize}
    \item What we do differently: A limited number of comics benchmarks exists in the literature
    \item What are the existing datasets, and what they are made for (briefly)
    \item general problems of datasets (availability, small, limited annotations)
    \item The only dataset with many annotations and varied is Manga109, but we also add ...
    \item see the table \ref{sample-table} where we show the composition of our benchmark dataset and the various element numbers.
\end{itemize}
\fi

A limited number of comics-related benchmarks exist in the literature, covering tasks such as classification (image classification \citep{chu_linebased_2014,daiku_comic_2018,jiang_learning_2018,terauchi_analysis_2019,xu_panelpageaware_2023}, emotion recognition \citep{nguyen_icdar_2021}, action detection \citep{iyyer_amazing_2017}), detection \citep{augereau_survey_2018}(panel, character, text, etc.), and modification (de-warping \citep{garai_automatic_2023}, image-to-image translation \citep{topal_dassdetector_2023}), to cite a few. We focus the discussion here on detection and analysis benchmarks and highlight the differences between \textit{CoMix} and prior work regarding the data collection process and available annotations and tasks.

One main limitation that affects all works related to comics (and art in general) is copyright issues. Among various datasets that have been proposed over the years, many no longer exist \citep{nguyen_comic_2017,wilber_bam_2017,inoue_crossdomain_2018,nguyen_comic_2017}, and many others are not available \citep{khan_color_2012,ho_redundant_2013,sun_specific_2013,qin_faster_2017,dunst_graphic_2017,he_endtoend_2018,cohn_visual_2023}. The majority of these datasets are designed for detection tasks, with images spanning from cartoon and sketches to French, American and Japanese comics. Only a few datasets are available whose annotations only assess classification \citep{guerin_ebdtheque_2013,iyyer_amazing_2017,nguyen_icdar_2021} and detection \citep{nguyen_digital_2018,fujimoto_manga109_2016,baek_coo_2022}.
% As discussed earlier, one main issue in comics is the absence of comprehensive and densely annotated datasets for low-level tasks such as object detection. As can be seen from Table \ref{tab:different-dataset-comparison}, the biggest datasets are not available, and, among the others, there is no consistency in annotations choices (i.e. if annotating all the characters, or only a few; if animals are characters or not, even if they ``talk''). 

Building on prior discussions, it's notable that existing datasets introduce sophisticated tasks such as speaker identification and re-identification; however, they predominantly feature manga-style comics and exhibit several critical shortcomings. Firstly, annotations are typically restricted to principal characters \citep{li_manga109dialog_2023}, limiting the scope of character detection and naming. Secondly, characters' names are often omitted \citep{sachdeva_manga_2024}, and thirdly, there are no established metrics or benchmarks for evaluating tasks like character naming and dialog generation. Furthermore, the Manga109 dataset \citep{li_manga109dialog_2023} is exclusively composed of Japanese mangas, which presents challenges for global applicability. Therefore, there is a clear need for a more inclusive comic dataset that not only spans multiple styles but also provides dense annotations and comprehensive metrics for benchmarking.

Moreover, data sources are of huge importance. \citet{sachdeva_manga_2024} selected the most popular manga of all time creating the PopManga dataset, gathering images from Manga Plus by Shueisha\footnote{\href{https://mangaplus.shueisha.co.jp/}{https://mangaplus.shueisha.co.jp}}. A similar approach was previously applied by \citet{iyyer_amazing_2017}, where a collection of 5k most-rated comic books was employed in constructing the COMICS, scraping images from the well-known Digital Comic Museum\footnote{\href{https://digitalcomicmuseum.com}{https://digitalcomicmuseum.com}}. These approaches ensure that data is of high quality but it does limit the variability in style and complexity. In fact, most common manga reflect user preferences, which force drawers to standardized style. The same, together with high-quality scans, happen to appear in comics. Moreover, as the sampled books correspond to the same collections, characters and styles for the PopManga appear to be the same, as in fact the PopManga unseen split is the collection of almost 1k images, but only spanning 10 different sagas.

Table \ref{tab:different-dataset-comparison} summarises the characteristics of the \textit{CoMix} compared to previous efforts. It can be observed that the \textit{CoMix} has better coverage of annotations. We emphasize that the \textit{CoMix} is not designed to be a large-scale dataset. Instead, it is an evaluation benchmark, with limited but densely annotated data, meant to assess the multi-task capabilities of models.

\begin{table}
  \caption{Characteristic of existing datasets (test split) compared to \textit{CoMix}. Tasks: Classification (c), Detection (d), Text-Character association (t2c), Character Re-Identification (c2c), Character Naming (N), and Dialog (D). {\color{ForestGreen} \cmark} represent available datasets, while {\color{red} \xmark} not, thus asterisks (\textbf{*}) represent reported numbers by authors. \colorbox{lightgreen!35}{Highlighted rows} are present in our benchmark dataset.}
  \vspace{3mm}
  \label{tab:different-dataset-comparison}
  \centering
  \begin{tabular}{lcccccrr}
    \toprule
    \textbf{Dataset} & \textbf{Release} &  \textbf{Avail} & \textbf{Tasks} & \textbf{Years} & \textbf{Style} & \textbf{Books} & \textbf{Pages} \\
    \midrule
    \rowcolor{lightgreen!35}
    eBDtheque \citep{guerin_ebdtheque_2013} & 2013 & {\color{ForestGreen} \cmark} & d,t2c & 1905-2012 & mix & 28 & 100 \\
    COMICS \citep{iyyer_amazing_2017} & 2017 & {\color{ForestGreen} \cmark} & c & 1938-1954 & comics & 3948 & 198k \\
    GCN \citep{dunst_graphic_2017} & 2017 & {\color{red} \xmark} & d,t2c & 1978-2013 & comics & *253 & *38k \\
    \rowcolor{lightgreen!35}
    DCM772 \citep{nguyen_digital_2018} & 2018 & {\color{ForestGreen} \cmark} & d & 1938-1954 & comics & 27 & 772 \\
    
    Manga109 \citep{fujimoto_manga109_2016,ogawa_object_2018} & 2018 & {\color{ForestGreen} \cmark} & d,t2c,c2c & 1970-2010 & manga & 109 & 10k \\
    BCBId \citep{dutta_bcbid_2022} & 2022 & {\color{ForestGreen} \cmark} & - & - & bangla & 64 & 3k \\
    VLRC & 2023 & {\color{red} \xmark} & - & 1940-now & - & *376 & *7k \\
    \rowcolor{lightgreen!35}
    PopManga \citep{sachdeva_manga_2024} & 2024 & {\color{ForestGreen} \cmark} & d,t2c,c2c & 2010-2023 & manga & 25 & 1.8k \\
    \midrule
    \rowcolor{lightgreen!35}
    % \textbf{Comics100} (our) & 2024 & {\color{ForestGreen} \cmark} & d,t2c,c2c,N,D & 1938-1954 & comics & 20 & 1.1k \\
    \textbf{\textit{CoMix}} (our) & 2024 & {\color{ForestGreen} \cmark} & d,t2c,c2c,N,D & 1938-2023 & mix & 100 & 3.8k \\
   
    \bottomrule
  \end{tabular}
\end{table}

\section{Books in the \textit{CoMix}}
\label{sec:books}
\iffalse
\begin{itemize}
    \item analysis of existing book datasets
    \item choice of books in comics100
\end{itemize}
\fi

% The \textit{CoMix} dataset is specifically designed to cover a vast diversity of comics styles books, hence inheriting samples from Japanese manga (PopManga), Americans (DCM and Comics100) and some French Bandes Desinè (eBDtheque). Multilingualism is not our main goal, so we do not include any Manga109 data, despite being well-designed and densely annotated. However, multilingualism occurs with some eBDtheque pages which are in French and Japanese. 
The \textit{CoMix} dataset has been meticulously curated to showcase a broad spectrum of comic book styles, drawing samples from Japanese manga (PopManga), American comics (DCM and newly collected Comics), and French Bandes Dessinées (eBDtheque). Although multilingualism is not a primary objective, some pages from eBDtheque include French and Japanese, reflecting incidental multilingual aspects.

% \begin{figure}
%     \centering
%     \includegraphics[height=10cm]{example-image-b}
%     \caption{Overview of character and book distribution for comics style dataset from DCM. We have selected a subset of 100 books that allows for maximum character variety across a fixed number of 100 books.}
%     \label{fig:comics-choise}
% \end{figure}

\textbf{Comics choice:} As depicted in Table \ref{tab:different-dataset-comparison}, the integration of existing datasets—PopManga, DCM, and eBDtheque—reveals a predominant bias towards manga-style comics. To address this imbalance, we strategically augmented the dataset with a selection of American comics from the Digital Comic Museum, which features over $22k$ golden-age American comic books. Popular characters such as \textit{``Plastic Man''} and \textit{``Daredevil''}, are drawn from the most downloaded comics, although selection criteria were refined beyond download counts due to potential skew from non-representative downloads \footnote{See comments in \href{https://digitalcomicmuseum.com/index.php?dlid=21158}{
Wanted Comics 11 -JVJ}}. Crucially, each book's metadata links to the Grand Comics Database \footnote{\href{https://www.comics.org/}{https://www.comics.org/}}, providing detailed character and storyline annotations that facilitated the selection process. 

Using this information, we selected comic books based on the distribution of character appearances across the books following the principle of (i) most characters possible, from the minimum number of books, and (ii) all possible instances of these characters should be in the 100 books selected, of which 20 goes to the test/val splits.
In constructing \textit{CoMix}, our primary goal is to underscore books featuring characters that recurrently appear across various publications, indicating significant narrative roles. To achieve this, the selection method computes the ratio of shared-to-unique character appearances for each book, thereby ranking and choosing the top books based on this ratio. Specifically, this ratio assesses the frequency of characters' appearances within the same book against their appearances in other books. To further refine our selection, the algorithm was designed to maximize the diversity of characters across the selected books, prioritizing the retention of books that uniquely feature these characters. This ensures a broad representation of characters, with most being exclusive to the selected subset and not featured in the broader set of $22k$ comics. The ratio is then exponentiated (in our implementation, squared) to enhance distinctions between books with high and low levels of character sharing. After calculating these ratios for all books, they are ranked in descending order based on their scores, leading to the selection of the top 100 books. This metric, referred to as the ``Pow Selection Approach'', is detailed in the Supplementary materials. % In Figure \ref{fig:comics-choice}, an overview of characters per book and character-pages visualization is provided. 
This approach ensures that the selected books reflect broader narrative arcs and character interconnectivity, thereby enriching the \textit{CoMix} dataset's utility for diverse comic book analysis.

\textbf{Splits:} The \textit{CoMix} contains 80 books from different sources, each of 44 pages on average, with a total of 3.5k images. This corresponds only to the held-out test split available through the evaluation server. Moreover, a validation split (20 books, 466 images), on which we tested and reported the results in this paper is provided.

\section{Annotation in the \textit{CoMix} benchmark}
\label{sec:anns}
\iffalse
\begin{itemize}
    \item Object detection
    \item Speaker identification
    \item Character re-identification
    \item Character naming
    \item Text-panel sorting
    \item Dialog generation
\end{itemize}
\fi
\begin{table}
  \caption{Annotations in the \textit{CoMix} benchmark. The reported numbers correspond to both test (80 books) and validation (20 books) splits. Each detected object is represented by the class and bounding box. Text elements also possess ground truth transcriptions. Character elements, whenever relevant, are associated with the name. Not all texts are associated with a character, but all the texts have Reading order ground truth.}
  \label{tab:dataset-stats}
  \centering
  \begin{tabular}{lr}
    \toprule
    \textbf{Annotation type} & \textbf{\# anns}  \\
    \midrule
    Object detection (4 classes) & 130k \\
    Speaker identification  & 29k \\
    Character Re-Id  & 59k \\
    Reading order  & 3.9k \\
    Character Naming  & 15k \\
    Dialog generation & 3.9k \\
    \bottomrule
  \end{tabular}
\end{table}

We annotate these comics with six types of annotations to cover low-level and high-level aspects, both computational and reasoning tasks. We enable various evaluations: object detection, speaker identification, character re-identification, character naming, reading order, and dialog generation. We include a summary of the number of annotations in Table \ref{tab:dataset-stats} and visualizations in Figure \ref{fig:first-figure}. % and the metrics used in section \ref{sec:baseline-results}.

\textbf{Object detection.} Object detection represents the root annotation of our benchmark. All the other annotations, except for dialog generation, are linked or grounded into detected objects. In the annotation process, we instructed annotators to focus on all size elements of the four selected categories: panels, characters, faces, and text boxes. Regarding other possible comic-related classes, we decided not to consider balloons and onomatopoeias, nor scene text not relevant to spoken text. Many of these classes were already annotated in the existing datasets, but with different design choices: PopManga does not have faces and panels, DCM does not have faces, and none of them have subsequent complex annotations that we describe next. In the supplementary material, we show our annotations compared to existing ones. Regarding characters, following \citep{guerin_ebdtheque_2013}, we consider human-like, animal-like, and object-like characters to be annotated, without explicitly differentiating these. When the characters are partially occluded, the annotators mark only the visible part of the character's body as boxes. Moreover, when the character is not recognizable if not thanks to in-page context (i.e. in a large zoom-out scene), we ask to annotate the so-called character. Some ambiguous character detection remains, like multi-page zoomout, or back-and-forth change of scene among pages. A list of specific examples is included in the supplementary materials. Finally, we consider faces where the nose is visible, independently from the face-side (front, side, or backward).

\textbf{Speaker identification.} Although object detection allows for some initial comics analysis, such as global comics layout with panels, character pose variability, and the usual aesthetic and location of text and dialogs, they do not fully describe a comic page. A better understanding of compositional interaction with objects arises by linking the text boxes (spoken texts) to the respective speaking characters. To this end, following \citep{li_manga109dialog_2023}, we annotate oriented polylines (from text to character) on the respective objects. Annotators were instructed to select two points, text, and characters, inside the respective boxes. These points are post-processed to obtain (text, character) pair indexes.

\textbf{Character Re-Identification.} Another linking task comprehends re-identify characters within the same page. This task can be seen as binary classification among all the possible pairs of characters (as approached in \citep{sachdeva_manga_2024}) or as a clustering task. When the characters are difficult to recall (i.e. recognizable only by looking at the context, even when reading the panels) we still ask the annotators to put in the effort and give the right identity to the character. These cases appear many times, especially in comics-style, as characters are often not consistent in little details (see example in the supplementary materials for a better overview).

\textbf{Reading order.} Mangas and Comics are read differently. Mangas from right to left, while Comics from left to right. Both are read top-down. Despite the good performances of cut-based panel sorting algorithms \citep{kovanen_layered_2015,hinami_fully_2021} and graph-based approach \citep{sachdeva_manga_2024}, we decided to extensively annotate also reading order to assess the corner-cases of these so-common used euristics.

\textbf{Character Naming.} Differently from the existing datasets \citep{fujimoto_manga109_2016}, which assign a name tag only to main characters, we extensively annotate names assigning to non-primary characters an exhaustive description of their role (i.e. ``Captain'', ``Sailor'') which ensure a non-zero metric score in character naming metric (see Section \ref{sec:baseline-results} for an overview of the metric). However, not every character can have a specific description, thus, we rely on incremental indexes to name different characters with the same roles (i.e. ``US soldier 1'', ``US soldier 2'', etc.). Whenever the character names are spoken in the text boxes, their names propagate on the whole story.

\textbf{Dialog Generation.} As a way to connect high-level and low-level single-page comics understanding capabilities, we define dialogs as an ordered list of tuples with the speaker name (character names or ``narrator''). The texts are sorted based on the ground truth reading order. Annotators are instructed to maintain punctuations and text format (lower-case and upper-case) but discard other text properties (bold, italics, canceled, or underlined text).

% \textbf{Annotation format.} Following \citep{fujimoto_manga109_2016}, we release XML files for detection, re-identification, text-panel sorting, and naming tasks. For the dialog, we provide ...

\section{Baselines results}
\label{sec:baseline-results}
\iffalse
\begin{itemize}
    \item The different complexity tasks are summarised in table \ref{sample-table-3}, where you can find the structure of the output, the metrics and the top-performing model with the score. As for detection, many of the models cannot detect all the objects, we report the total metric (considering as zero when the model cannot detect the object) and the filtered scores (considering only detectable objects scores).
    \item baselines: ideally, a single model should be able to perform all the task, but such a model is not available in the literature, we include results obtained with per-task baselines. The most general model is Magi, which performs 3 over the 6 tasks, and another task partially (dialog with no names). results are shown in table \ref{sample-table-3}.
\end{itemize}
\fi

\begin{table}
  \caption{Computational tasks, and top-performing baselines in the \textit{CoMix}. The dialog proposed metric ``Hybrid Dialog Score'' is indicated with HDS.}
  \vspace{3mm}
  \label{tab:metric-results}
  \centering
  \begin{tabular}{lcccr}
    \toprule
    \textbf{Task} & \textbf{Output} & \textbf{Metric} & \textbf{Baseline} & \textbf{Score} \\
    \midrule
    Object detection & box detection & mAP - $R@$100 & Magi  & 78.6 - 67.9\\
    Speaker identification & object indexes & $R@$\#text & heuristic & 0.68 \\
    Character Re-Id & cluster ids & AMI - NMI & DINOv2 & 0.29 - 0.51 \\
    % Reading order & sorted text id & edit distance & Magi \citep{sachdeva_manga_2024} & {\color{red} ???} \\
    Character Naming & names & ANLS & GPT-4  & 47.11 \\
    Dialog generation & list of tuples & HDS & GPT-4  & 93.14 \\
    \bottomrule
  \end{tabular}
\end{table}

\textbf{Computational tasks:} We defined six computational tasks based on the annotations available in the \textit{CoMix} dataset. We summarise in Table \ref{tab:metric-results} the task definitions (outputs, metrics) and the performance of top-performing baselines. It can be observed that the \textit{CoMix} combines lower-level dense prediction tasks like object, speaker identification, and character re-identification, whose outputs are box and group of indexes, with higher-level tasks like dialog generation. More details about the task definitions are included in the supplementary materials.

\textbf{Baselines:} Ideally, a single model should be able to perform all the tasks in the \textit{CoMix} benchmark. Since such a model is not available in the literature, we include results obtained with per-task baselines on the validation split for all the six tasks in the \textit{CoMix}; see Table \ref{tab:metric-results} for a summary of top-performing baselines and their average performance, and the supplementary materials for more details. When selecting and running these baselines, we favored the same approaches used by \citet{sachdeva_manga_2024}, which we carefully detail in the following section. However, for character naming, such models do not exist in the literature, so we evaluated namings together with dialogs, instead of with detection and clustering. 

In the following section, we provide, for every benchmark, an explanation of the used metrics and an overview of the methods employed in the benchmark.

\textbf{Object detection:} For object detection, we selected two common metrics: mean Average Precision at IoU of $0.5$ ($mAP@0.5$) and Recall at 100 objects ($R@100$). We benchmarked a variety of models including convolutional and transformer-based architectures, with a focus on their performance in both fine-tuned and zero-shot settings on comic data. Among these, GroundingDino \citep{liu_grounding_2023}, designed for open-set object detection using natural language object classes, was a key zero-shot model utilized to detect four classes: panels, characters, text, and faces. Additionally, conventional models such as Faster R-CNN, SSD, and YOLO were trained on varied comic styles to assess the impact of different training data distributions. For characters and faces detection we employed also a YOLOX-based available model named DASS \cite{topal_dassdetector_2023}. Notably, Magi \citep{sachdeva_manga_2024}, a transformer-based model, was included for its impressive capability demonstrated in Manga-style comics. Details about the fine-tuned training procedure and GroundingDino prompts for zero-shot detection are provided in the appendix, as well as more detailed benchmark results.

\textbf{Text-Character association.} The task of associating a speaking character to a textbox is not new, and neither is the metric. Following previous work \citep{li_manga109dialog_2023}, we employ classical Recall@K, with the value of K indicating the number of textboxes on a page, called $R@\text{\#text}$. The global score is obtained by averaging a single-page score. The benchmarks are composed by: (i) an existing heuristic approach connecting the textbox to the closest character \citep{nguyen_digital_2018}; and (ii) Magi, which is trained for providing these associations among the detected textboxes and characters. In this case, differently from what is shown by \citet{sachdeva_manga_2024}, the heuristic of ``most close character'' works better. We speculate this drop in performance is given by the style shift between manga (on which Magi has been trained) and comics.

\textbf{Character Re-Identification.} For the task of Re-Identification (also known as single-page character Clustering), we employed Adjusted Mutual Information ($AMI$) and Normalized Mutual Information ($NMI$) to evaluate performance. $AMI$ measures the agreement between the clustering results and the ground truth labels, adjusted for chance. $NMI$, on the other hand, normalizes mutual information by the potential disorder in each set of labels, thus reflecting the purity of the clustering. Contrary to previous other studies, we did not utilize retrieval metrics such as $MRR$, $MAP$, or $Precision@k$. These metrics are heavily dependent on the presence of relevant items within the retrieved sets. Given that our analysis involves clustering where unequivocally relevant retrievals are absent — often resulting in clusters of a single element — these metrics would inherently score zero, thus failing to provide meaningful insights into our clustering approach's effectiveness. We benchmarked CLIP and DINOv2 as feature extraction models calculating the best clusters at max $AMI$ score, following \citep{sachdeva_manga_2024}. Differently from what was previously reported, DINOv2 obtained higher scores than the fine-tuned model competitor Magi, thus indicating that despite maintaining high detection scores, Magi is not able to retain recognition performances out of its manga domain.

\textbf{Reading order.} For the task of reading order, we propose a simple edit distance metric on the sorted detected textboxes matched with the ground truth. As Magi performances in detecting textboxes and panels retain high accuracy, as benchmark adapted the Magi algorithm to operate both for manga and comics style. Prior knowledge makes it possible to know that comics pages are read from top to bottom, with an orientation difference between manga (right to left) and comics (left to right). Once the detected panels are sorted (with DAG approaches described in \citep{sachdeva_manga_2024}, adapted for comics), the textboxes within the panels are ordered based on the vicinity with the panel's top-right corner (manga) or top-left corner (comics). The result is reported in Table \ref{tab:metric-results}, and an overview of the panel DAG is given in the supplementary materials.

\textbf{Character Naming \& Dialog Generation.} In addressing the challenges of character naming and dialog generation, we acknowledge that there is a lack of specific metrics. In previous works, the dialog is generated with a combination of algorithmic panel ordering, character progressive naming (``character 1'', ``character 2''), and fine-tuned OCR for textbox transcription. This engineering and multi-step approach is a good showcase but is not evaluated with some metric against any ground truth. We introduce a smoothed case-sensitive edit distance metric called ``Hybrid Dialog Score'' ($HDS$) to evaluate the accuracy of model outputs against ground truth dialog annotations. This metric assesses both the precision of transcribed dialogues and the accuracy of character identification in a unified framework. The metric operates in three steps: (i) we match ground truth and predictions texts using Hungarian matching with edit distance metric; for every match, we calculate (ii) the textbox edit distance (normalized on ground truth text length) and (iii) the character name ANLS. An in-depth overview of the metric is provided in the supplementary materials, together with pseudocode. For our benchmarks, we utilized Magi as described in  \citep{sachdeva_manga_2024}. As MAGi is not able to detect names, we employ GPT-4 \citep{openai_gpt4_2024}, leveraging its multimodal capabilities to interpret comic pages and generate structured outputs that include both dialogues and character names. GPT-4 was tasked with identifying and naming characters, where known, or assigning incremental identifiers otherwise, and transcribing dialogues exactly as they appear in comic format, respecting the sequence and case sensitivity. This approach not only captures the complexity of comic narratives but also enhances the evaluation of dialog transcription fidelity and character consistency across various comic styles. The result is reported in Table \ref{tab:metric-results}, and an overview of the GPT-4 prompt to obtain structured predictions is given in supplementary materials.

\section{Ethical Considerations}
\label{sec:ethical}

\paragraph{Copyrights and Consent.} Comics, as a form of artistic expression, are governed by copyright laws that restrict access and usage. Our dataset, \textit{CoMix}, aggregates comics from diverse sources: American comics from the Digital Comic Museum, manga from PopManga, and French Bande Dessinée from eBDtheque. In particular, \textit{Digital Comic Museum} contains public domain assets, either released without copyright or with expired rights, allowing free research use \cite{lund2015mutant}. The \textit{PopManga} images are publicly accessible on ``Manga Plus by Shueisha'' with official permissions from copyright holders. The \textit{eBDtheque} provides publicly available data cleared for non-commercial research use.

To ensure compliance, we have reorganized the dataset's structure, enabling users to acquire images directly from original sources and utilize our repository tools for formatting and validation \citep{Vivoli2024ComicsDF}. This approach maintains adherence to copyright norms while promoting dataset accessibility and replicability.

\paragraph{Data Quality and Representativeness.} The \textit{CoMix} benchmark evaluates models across various comic styles—American, Japanese, and European—to represent major comic production hubs. Predominantly comprising out-of-copyright works, especially American comics from the 1950s, the dataset may inherently reflect the social biases and stereotypes of that era. To address potential biases, we ensure a balanced representation of comic styles and origins to minimize cultural bias (Diverse Dataset Composition) and incorporate statistical analysis of factors like gender, ethnicity, and language representation (Bias Detection and Analysis).

On this regard, our first analysis concerns biases related to appearance (color vs. black-and-white) and types of characters in the comics. The color statistics are straightforward: by examining the first pages, we could determine whether each comic chapter is in color or black-and-white, as shown in Table \ref{tab:wab}. For character type analysis, we classified character crops into one of four categories listed in Table \ref{tab:char}, using a state-of-the-art open-source MLLM\footnote{MiniCPM-llama3-v-2.5 at \url{https://huggingface.co/openbmb/MiniCPM-Llama3-V-2\_5}}), which is known for its strong visual recognition and instruction following capabilities.

While the dataset shows a bias towards male characters, this is considered a reflection of actual character type distributions in real comics and manga, rather than a flaw in the dataset's quality. This dataset is intended for tasks such as dialog transcription testing.

\begin{table}[h]
    \centering
    \caption{Initial Statistics in the \textit{CoMix} Dataset}
    \begin{subtable}[b]{0.45\linewidth}
        \centering
        \caption{Color vs. black-and-white images.}
        \label{tab:wab}
        \begin{tabular}{c|c}
            \toprule
            \textbf{Type} & \textbf{Percentage} \\
            \midrule
            Color & 59.2\% \\
            Black-and-White & 40.8\% \\
            \bottomrule
        \end{tabular}
    \end{subtable}
    \hfill
    \begin{subtable}[b]{0.45\linewidth}
        \centering
        \caption{Character Types in the \textit{CoMix} Dataset}
        \label{tab:char}
        \begin{tabular}{c|c}
            \toprule
            \textbf{Character Type} & \textbf{Percentage} \\
            \midrule
            Male & 74.3\% \\
            Female & 17.9\% \\
            Animals & 6.3\% \\
            Other & 1.5\% \\
            \bottomrule
        \end{tabular}
    \end{subtable}
\end{table}

Future iterations will expand to include underrepresented styles such as Webtoons and Manhwa, alongside additional languages, to enhance global applicability and reduce cultural over-representation. Detailed statistics and bias analyses are provided in the supplementary materials.

\paragraph{Semantic Harmful Content.} Our dataset has been meticulously curated to exclude NSFW and offensive content. All included works comply with the Comics Code Authority guidelines, ensuring appropriateness for diverse research and educational purposes. We conducted an automatic analysis of semantic content using the Llama3-80B model to classify panel-level text for offensiveness. Initial findings indicate low levels of harmful content (less than 0.5\%), with detailed results and model prompts available in the supplementary materials.

Overall, these ethical considerations ensure that our research adheres to high standards of integrity, fairness, and respect for diverse cultural norms \cite{lund2015mutant}.

\section{Conclusion}
\label{sec:conclusion}

We introduce \textit{CoMix}, a novel benchmark for multi-task and multi-modal comic analysis that addresses the limitations of existing datasets by incorporating diverse comic styles—including American, manga, and French—and providing comprehensive annotations across a wide range of tasks. \textit{CoMix} encompasses fundamental vision tasks such as object detection and character re-identification, alongside complex multi-modal reasoning tasks like character naming and dialogue generation. The introduction of the Hybrid Dialog Score offers innovative metrics for evaluating these advanced tasks.

Baseline evaluations reveal significant performance gaps between state-of-the-art models and human performance, highlighting the inherent challenges in achieving nuanced understanding of the interplay between visual and textual elements in comics. By releasing the \textit{CoMix} validation split and establishing an evaluation server for the held-out test split, we promote open research and facilitate robust benchmarking. \textit{CoMix} sets a new standard for comprehensive comic analysis, providing a diverse and challenging testbed that will drive the development of more sophisticated and generalizable models capable of human-like comprehension in this culturally rich medium.

\newpage

\begin{ack}
This paper has been supported by the Consolidated Research Group 2021 SGR 01559 from the Research and University Department of the Catalan Government, and by project PID2023-146426NB-100 funded by MCIU/AEI/10.13039/501100011033 and FSE+.
This work has been also funded by the European Lighthouse on Safe and Secure AI (ELSA) from the European Union’s Horizon Europe program under grant agreement No 101070617.

Moreover, we are grateful to Digital Comics Museum for providing a free accessible source of comic books and to Grand Comics Database for the effort of building and maintaining correct and updated versions of comic memories. We want to thank Mohammed Ali Souibgui and Marco Mistretta for providing insightful input and Niccolò Biondi, Irene Campaioli, and Mariateresa Nardoni for being part of the revision and annotation team.
\end{ack}

% \section*{References}

\bibliography{main}

\newpage

\clearpage % Start supplementary material on a new page

% Reset Section Numbering for Supplementary
\setcounter{section}{0}
\renewcommand{\thesection}{S\arabic{section}}

% Supplementary Title Block

\suptitleblock

This supplementary document complements the main paper by providing additional information and examples that could not be included within the page constraints of the original manuscript. The structure of this document is as follows: % in Section \ref{algo:pow-selection} we describe the book selection while 
in Section \ref{sec:fine-tuned-details} details about the models and their usage. In Section \ref{sec:anns-overview} we presents an overview of the annotations from existing datasets and compares them with those in the \textit{CoMix} benchmark. In Section \ref{sec:data-selection} we detail the selection process of American Comics books included in our study. Extensive results are provided in Section\ \ref{sec:results}, including those reported in the main manuscript. Finally, copyright and biases information is outlined in the Ethical Section \ref{sec:add-ethical}.
This supplementary material is intended to enhance the understanding and transparency of the research presented in the main paper.

\section{Detailed Model Descriptions and Settings}
\label{sec:fine-tuned-details}
\textbf{GroundingDino:} For zero-shot detection, GroundingDino was pivotal, using an array of class prompts to adapt to comic-specific elements. The class prompts used for detection were:
\begin{itemize}
    \item \textbf{Panels}: "comics panels", "manga panels", "frames", "windows"
    \item \textbf{Characters}: "characters", "comics characters", "person", "girl", "woman", "man", "animal"
    \item \textbf{Text}: "text box", "text", "handwriting"
    \item \textbf{Faces}: "face", "character face", "animal face", "head", "face with nose and mouth", "person's face"
\end{itemize}
These prompts enabled the model to flexibly identify and classify a wide range of comic book elements by interpreting each class through the lens of natural language descriptions.

\textbf{DASS:} A convolution-based model utilizing the YOLOX architecture, DASS was developed in a self-supervised setup using a distillation approach from a teacher network with OHEM loss. It includes three variants—DCM, manga109, and mix—each fine-tuned on the dataset reflecting its name, optimized for detecting styles consistent with its training data.

\textbf{Standard Models:} Faster R-CNN, SSD, and YOLO models were adapted for comics by training them on comics, manga, and mixed datasets. These models were initialized with standard configurations and then fine-tuned to tailor to comic data, with adjustments such as changing the output classes to four and modifying learning rates and decay settings to optimize performance. Specifically, Faster R-CNN was adapted using a ResNet-50 backbone and trained with a learning rate of $5e^{-3}$, along with employing both StepLR and CosineAnnealingLR schedulers to manage learning rate adjustments across epochs. YOLOv8 and SSD have been trained using default configurations from ``ultralytics'' \footnote{\href{https://github.com/ultralytics/ultralytics}{https://github.com/ultralytics/ultralytics}} and ``mmdetection''\footnote{\href{https://github.com/open-mmlab/mmdetection}{https://github.com/open-mmlab/mmdetection}} frameworks, respectively.

\textbf{Magi:} The transformer-based Magi model, following RelationFormer architecture \citep{shit_relationformer_2022}, integrates a DeTr backbone with two MLP heads, focused on speaker identification and character re-identification. Initially pre-trained on a noisy dataset from Mangadex annotated with GroundingDino and later fine-tuned on a specialized popmanga dev-set, Magi exemplifies advanced model training with a focus on specific comic interactions.

%%%%%%%%%%%%%%%%%%%%%%%%%%%%%%%%%%%%%%%%%%%%%%%%%%%%%%%%%%%%

\section{Annotations overview}
\label{sec:anns-overview}
\begin{figure}[h]
\centering
\begin{subfigure}{.48\textwidth}
  \centering
  \includegraphics[width=0.9\linewidth]{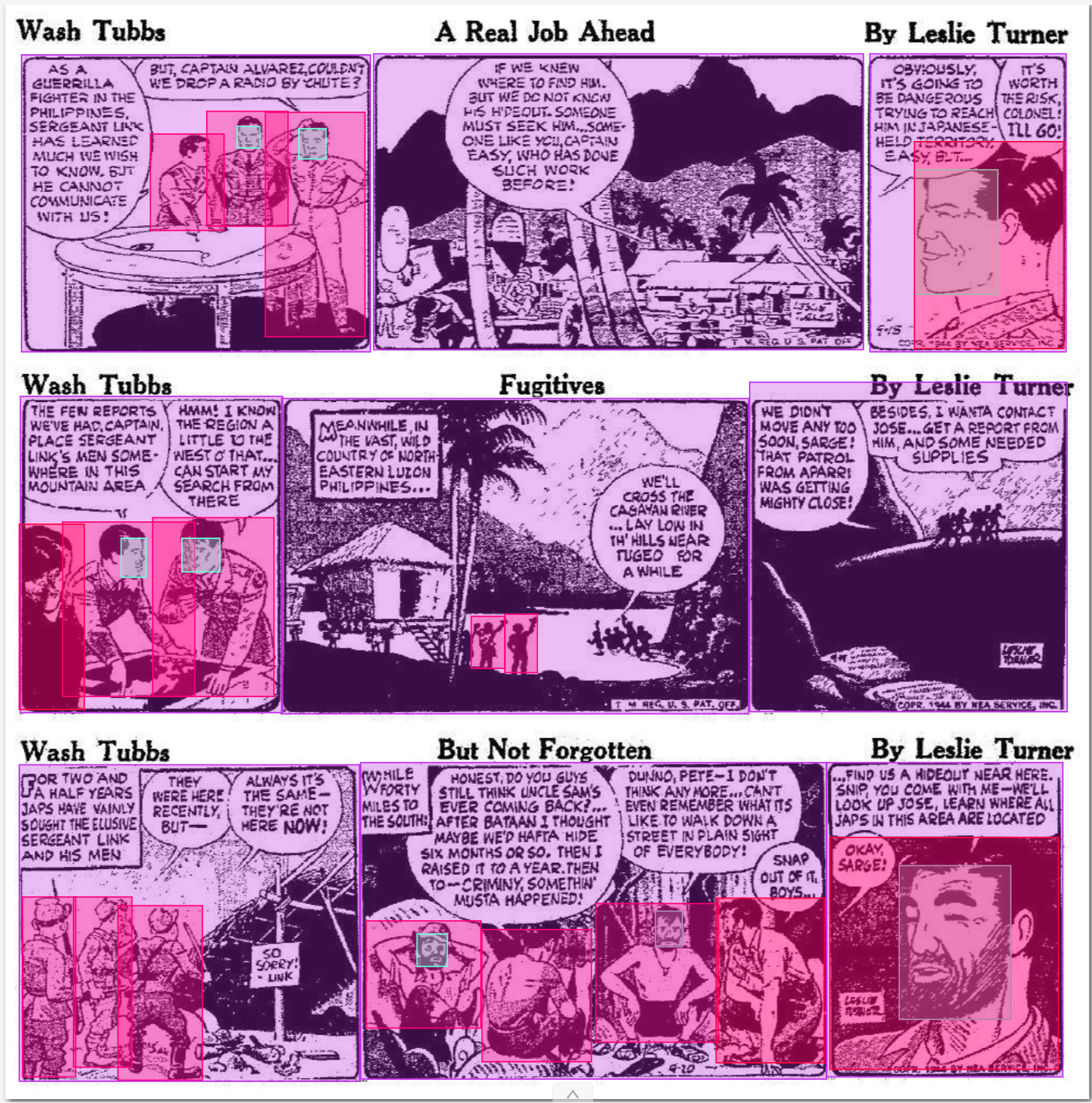}
\end{subfigure}%
\begin{subfigure}{.48\textwidth}
  \centering
  \includegraphics[width=0.9\linewidth]{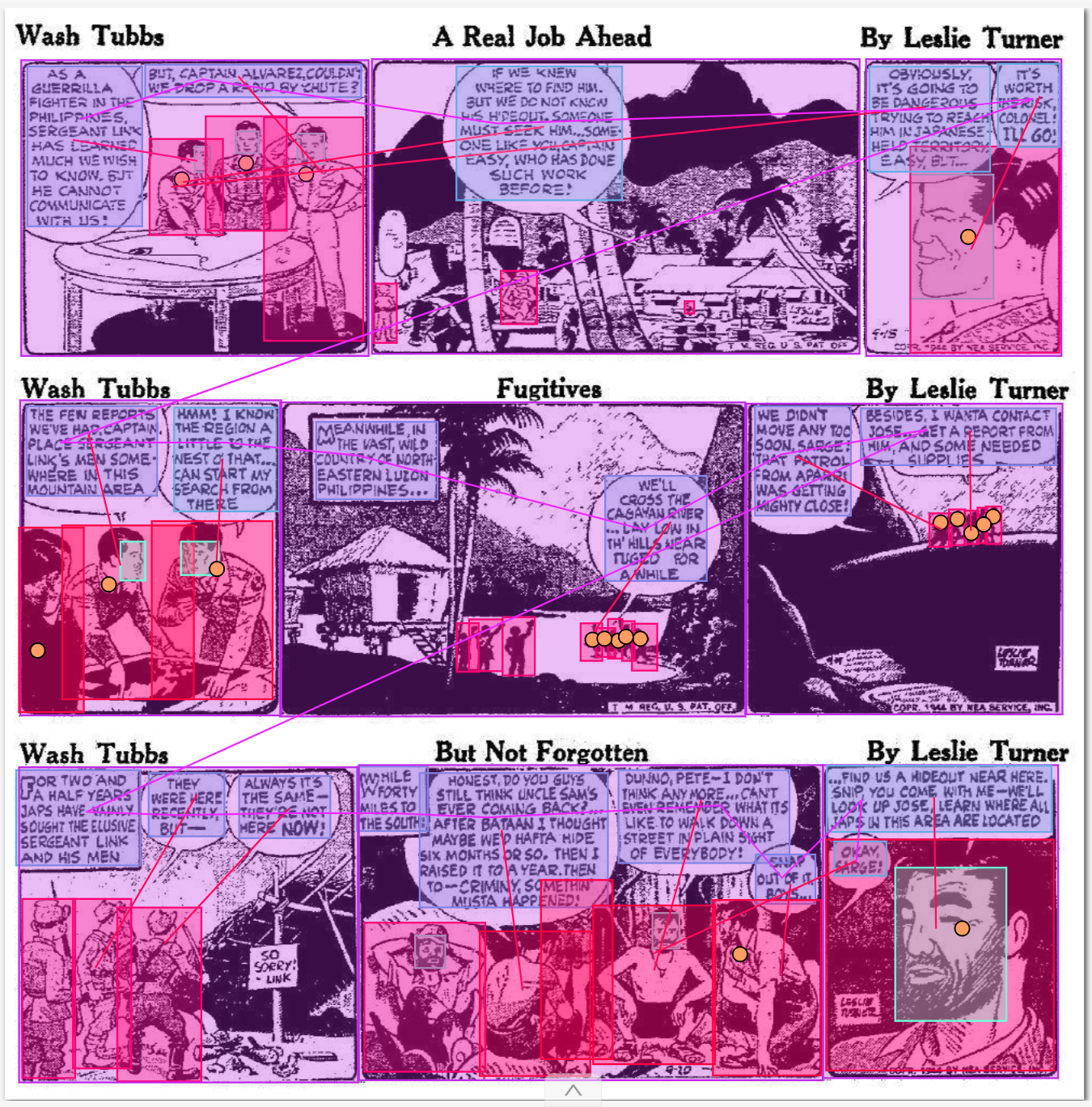}
\end{subfigure}
\caption{Image from ``DCM'', original annotations (left) and our \textit{CoMix} corrected and integrated annotations (right). Every point indicates a re-identified character.}
\label{fig:dcm-before-after}
\end{figure}

\begin{figure}[h]
\centering
\begin{subfigure}{.48\textwidth}
  \centering
  \includegraphics[width=0.9\linewidth]{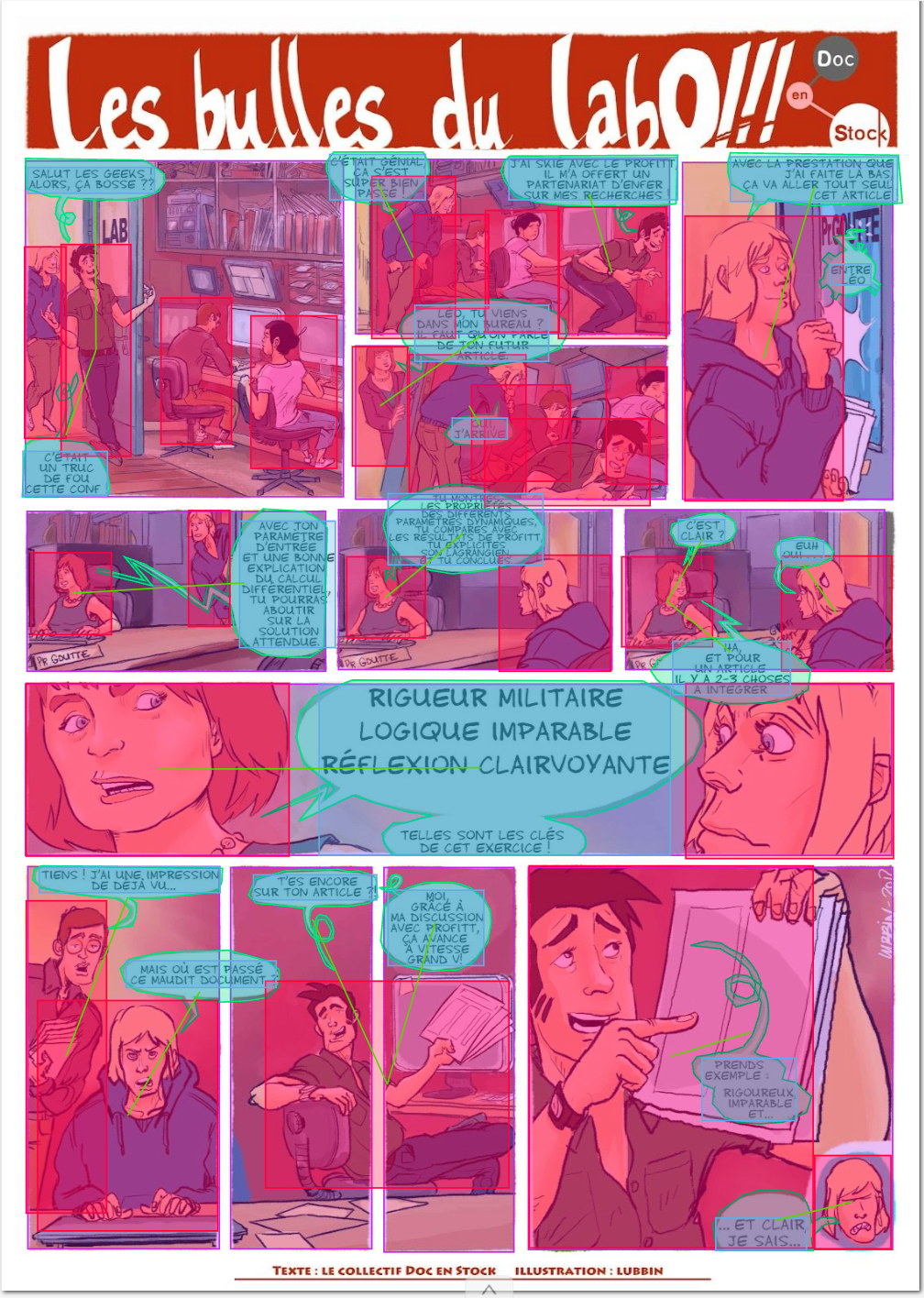}
\end{subfigure}%
\begin{subfigure}{.48\textwidth}
  \centering
  \includegraphics[width=0.9\linewidth]{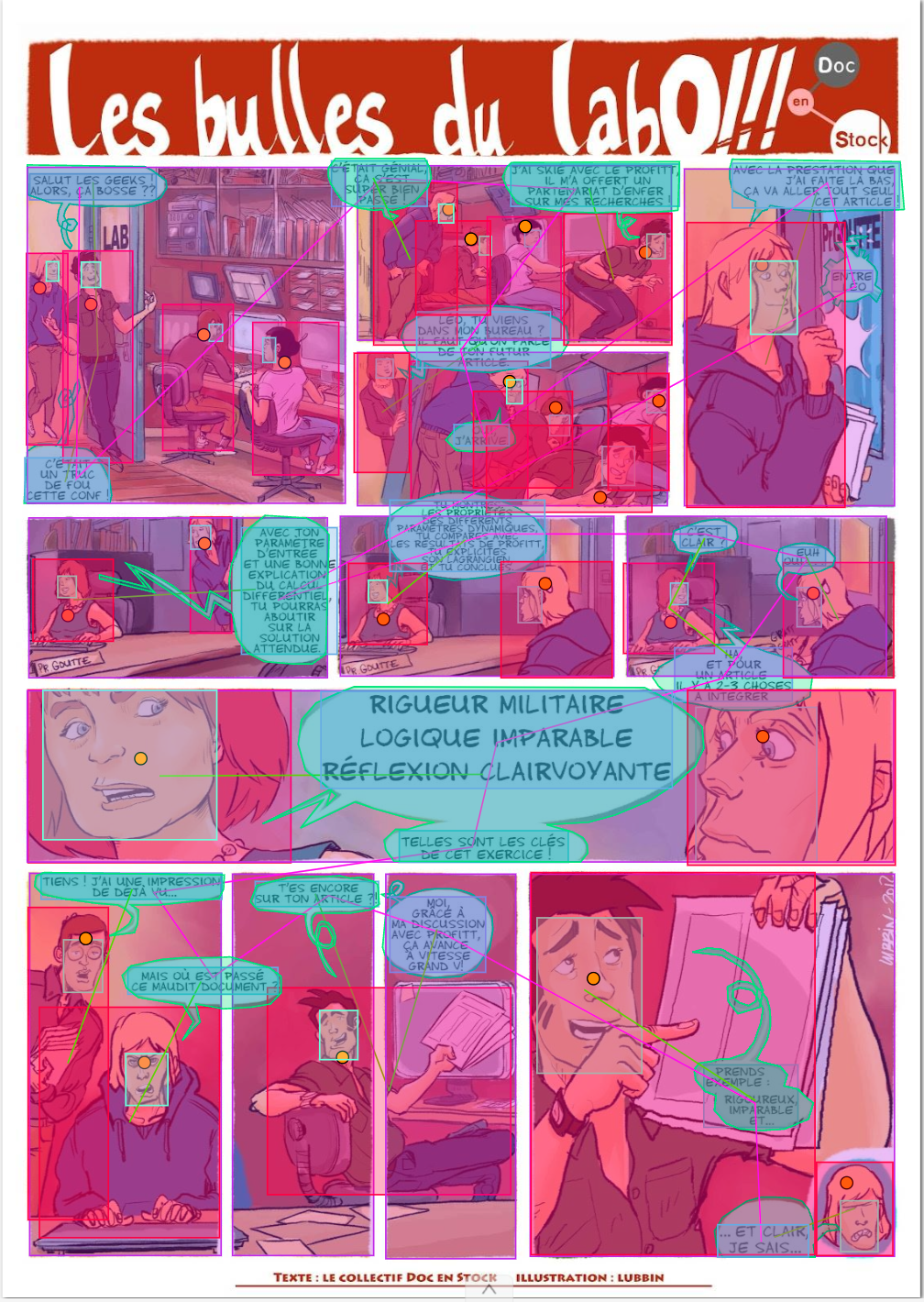}
\end{subfigure}
\caption{Image from ``eBDtheque'', original annotations (left) and our \textit{CoMix} corrected and integrated annotations (right).}
\label{fig:ebd-before-after}
\end{figure}

\begin{figure}[h]
\centering
\begin{subfigure}{.48\textwidth}
  \centering
  \includegraphics[width=0.9\linewidth]{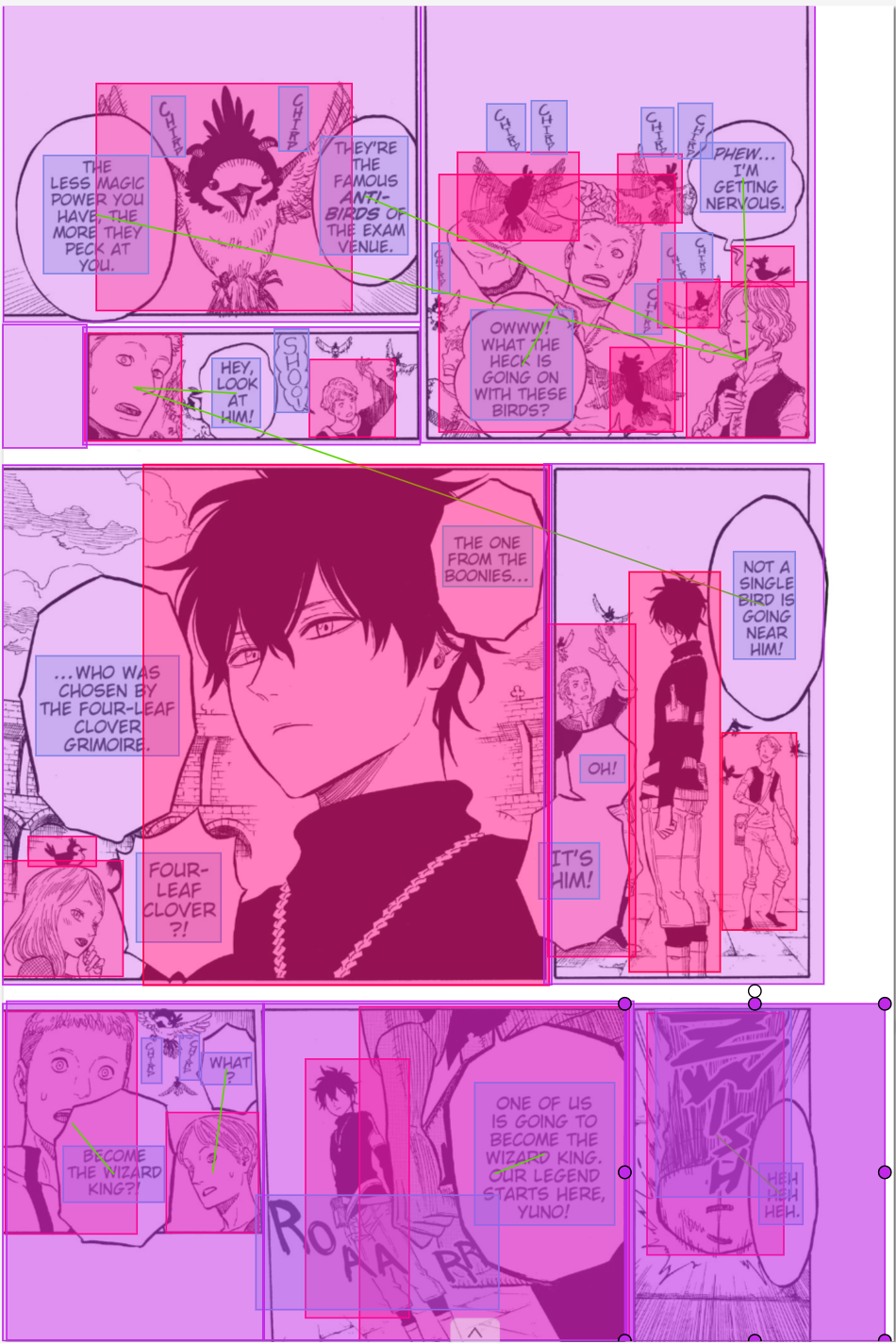}
\end{subfigure}%
\begin{subfigure}{.48\textwidth}
  \centering
  \includegraphics[width=0.9\linewidth]{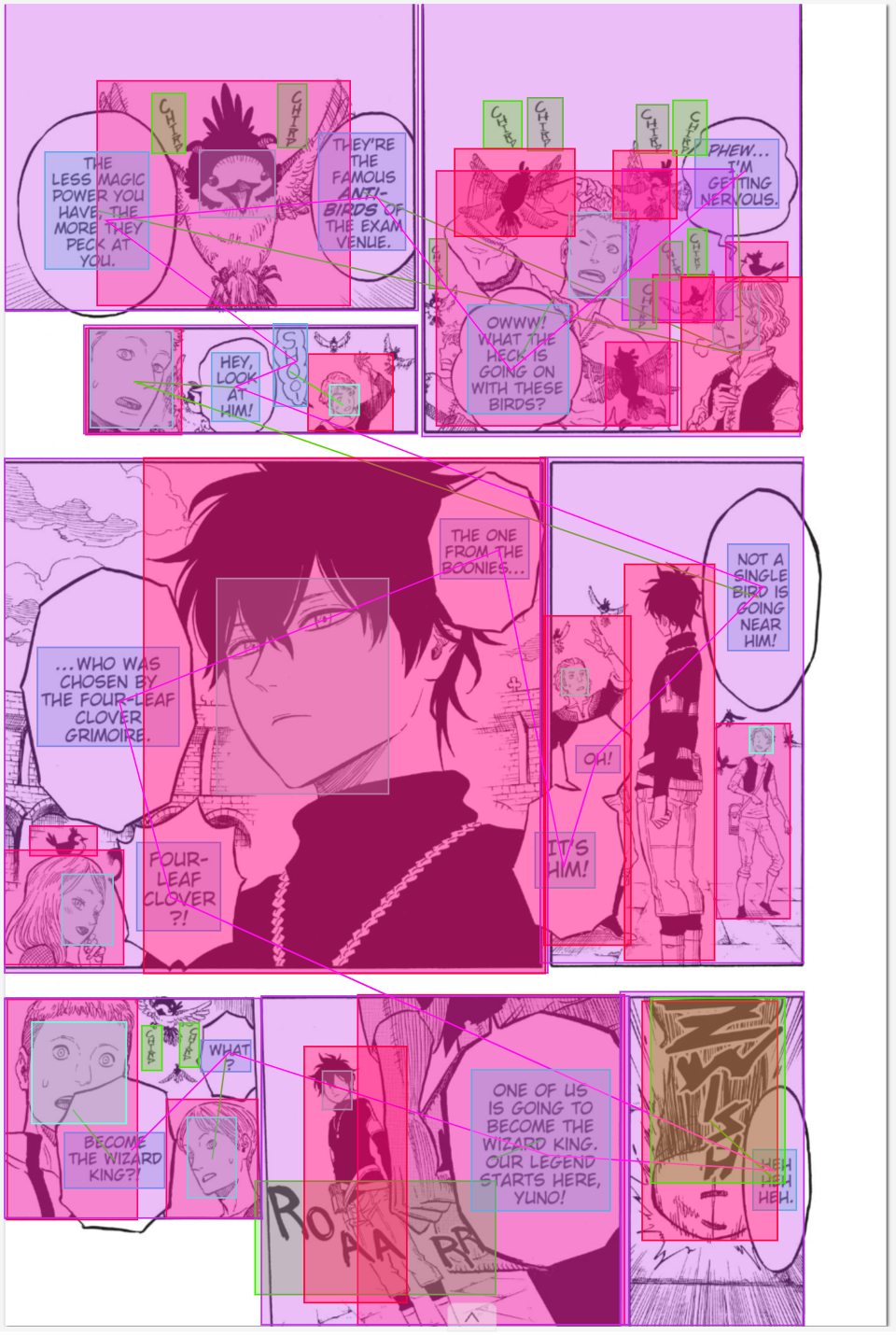}
\end{subfigure}
\caption{Image from ``PopManga'', original annotations (left) and our \textit{CoMix} corrected and integrated annotations (right).}
\label{fig:pop-before-after}
\end{figure}

This section offers an overview of the annotation differences in the \textit{CoMix} benchmark compared to previous standards. Specifically, distinctions are highlighted in Figures \ref{fig:dcm-before-after}, \ref{fig:ebd-before-after}, and \ref{fig:pop-before-after}, where the "before" annotations are displayed on the left and the "after" on the right.

We state in the main manuscript that our benchmark does not include detection of objects such as "Balloons", "Onomatopoeias", and "scene text" (non-spoken text boxes). This decision is based on two main considerations:
\begin{itemize}
\item[\textbf{(i)}] \textbf{Balloons:} Often, the annotated textboxes (spoken texts) are located inside balloons, making balloons essentially wrappers for the text. However, sometimes the text appears in a narrative box without a contour, making balloon detection a weak approximation.
\item[\textbf{(ii)}] \textbf{Onomatopoeias and Scene Text:} If the text is spoken by a character, it is detected as "textbox", different from other approaches such as \cite{fujimoto_manga109_2016}. When onomatopoeias illustrate sound effects, we omit them to focus exclusively on spoken text, aligning with \cite{li_manga109dialog_2023}. Unlike \cite{sachdeva_manga_2024}, we do not annotate scene-text to ensure models focus on crucial text (spoken) and avoid distractions from the scene.
\end{itemize}

Despite eBDtheque offering balloon annotations as seen in Figure \ref{fig:ebd-before-after}, we chose not to eliminate these annotations. Similarly, PopManga, shown in Figure \ref{fig:pop-before-after}, provides bounding boxes for all text present in the scene, including spoken text, onomatopoeias, or scene text. We differentiate these by introducing onomatopoeia and scene-text classes without utilizing them in our analyses. These annotations are preserved for potential future research.

Additionally, annotations for speaker identification, character re-identification, and reading order may be less visible in the images provided. These are represented as green and red lines connecting the text to the character boxes (speaker identification), colored points identifying each character bounding box (character re-identification), and purple splines connecting the text boxes in reading order orientation (reading order). For a clearer view of these annotations, we will release the validation set images and annotations on the project website\footnote{Repository link: \url{https://github.com/emanuelevivoli/CoMix-dataset}.}.

High-level annotations such as character naming and dialogue generation are not shown in these images but are part of our detailed annotation framework.

\subsection{Quantitative details}
In Table \ref{tab:dataset_summary-det} and Table \ref{tab:dataset_summary-high} are provided the original annotations numbers and the one we provide in \textit{CoMix}, comparing images from the same sources. The first row, of every annotation category, provides the existing annotation number, while the second row is the one in \textit{CoMix}. A third row provides the difference in percentage. From Table \ref{tab:dataset_summary-det} we can notice that almost all the annotations categories experienced a substantial increase (with some also 100\%) apart from the Panel detection in eBDtheque that we fixed eliminating duplicates and redundant panels boxes. For Table \ref{tab:dataset_summary-high}, almost all annotations in \textit{CoMix} were not present before.

\begin{table}[h]
\footnotesize
\centering
\begin{minipage}[t]{0.48\textwidth}
    \centering
    \caption{Summary of the Detection annotations across different datasets}
    \vspace{3mm}
    \label{tab:dataset_summary-det}
    \begin{tabular}{@{}lcccc@{}}
    \toprule
    \textbf{Category} & \multicolumn{4}{c}{\textbf{Data Type}} \\
    \cmidrule(l){2-5} 
                      & \textbf{DCM} & \textbf{EBD} & \textbf{Comics} & \textbf{Pop} \\ 
    \midrule
    \multirow{2}{*}{\textbf{Panel}}            & 4.5k  & 0.85k  & -     & -     \\
                                               & 4.6k  & 0.84k  & 6.7k  & 9.9k  \\
                                               & \tiny\textcolor{forestgreen}{+2.22\%} & \tiny\textcolor{forestred}{-1.18\%} & \tiny\textcolor{forestgreen}{+100\%} & \tiny\textcolor{forestgreen}{+100\%} \\
    \cmidrule(l){2-5} 
    \multirow{2}{*}{\textbf{Character}}        & 10.8k & 1.6k   & -     & 18.8k \\
                                               & 11.4k & 2k     & 15.9k & 19.5k \\
                                               & \tiny\textcolor{forestgreen}{+5.56\%} & \tiny\textcolor{forestgreen}{+25.00\%} & \tiny\textcolor{forestgreen}{+100\%} & \tiny\textcolor{forestgreen}{+3.72\%} \\
    \cmidrule(l){2-5} 
    \multirow{2}{*}{\textbf{Text}}             & -     & 1.1k   & -     & 20.8k \\
                                               & 8.4k  & 1.1k   & 11.9k & 16.5k \\
                                               & \tiny\textcolor{forestgreen}{+100\%} & \tiny\textcolor{lightgrey}{equal} & \tiny\textcolor{forestgreen}{+100\%} & \tiny\textcolor{forestgreen}{-20.67\%} \\
    \cmidrule(l){2-5} 
    \multirow{2}{*}{\textbf{Face}}             & 5.4k  & -      & -     & -     \\
                                               & 5.5k  & 1.1k   & 12.5k & 13.6k \\
                                               & \tiny\textcolor{forestgreen}{+1.85\%} & \tiny\textcolor{forestgreen}{+100\%} & \tiny\textcolor{forestgreen}{+100\%} & \tiny\textcolor{forestgreen}{+100\%} \\
    % \midrule
    % \textbf{Books}                            & 27    & 28     & 15    & 15    \\
    % \textbf{Images}                           & 0.76k & 0.08k  & 0.93k & 1.78k \\
    \bottomrule
    \end{tabular}
\end{minipage}\hfill
\begin{minipage}[t]{0.48\textwidth}
    \centering
    \caption{Summary of Higher-Level annotations across different datasets}
    \vspace{3mm}
    \label{tab:dataset_summary-high}
    \begin{tabular}{@{}lcccc@{}}
    \toprule
    \textbf{Category} & \multicolumn{4}{c}{\textbf{Data Type}} \\
    \cmidrule(l){2-5} 
                      & \textbf{DCM} & \textbf{EBD} & \textbf{Comics} & \textbf{Pop} \\ 
    \midrule
    \multirow{2}{*}{\textbf{Speaker ID}}       & -     & -      & -     & 13.6k \\
                                               & 6.2k  & 0.9k   & 8.9k  & 13.7k \\
                                               & \tiny\textcolor{forestgreen}{+100\%} & \tiny\textcolor{forestgreen}{+100\%} & \tiny\textcolor{forestgreen}{+100\%} & \tiny\textcolor{forestgreen}{+0.74\%} \\
    \cmidrule(l){2-5} 
    \multirow{2}{*}{\textbf{Character Re-ID}}  & -     & -      & -     & 15.8k \\
                                               & 7.4k  & 1.5k   & 8.5k  & 15.8k \\
                                               & \tiny\textcolor{forestgreen}{+100\%} & \tiny\textcolor{forestgreen}{+100\%} & \tiny\textcolor{forestgreen}{+100\%} & \tiny\textcolor{lightgrey}{equal} \\
    \cmidrule(l){2-5} 
    \multirow{2}{*}{\textbf{Reading Order}}    & -     & -      & -     & -     \\
                                               & 8.4k  & 1.1k   & 11.9k & 16.5k \\
                                               & \tiny\textcolor{forestgreen}{+100\%} & \tiny\textcolor{forestgreen}{+100\%} & \tiny\textcolor{forestgreen}{+100\%} & \tiny\textcolor{forestgreen}{+100\%} \\
    \cmidrule(l){2-5} 
    \multirow{2}{*}{\textbf{Character Naming}} & -     & -      & -     & -     \\
                                               & 4.4k  & 0.5k   & 6k    & 4.7k  \\
                                               & \tiny\textcolor{forestgreen}{+100\%} & \tiny\textcolor{forestgreen}{+100\%} & \tiny\textcolor{forestgreen}{+100\%} & \tiny\textcolor{forestgreen}{+100\%} \\
    \cmidrule(l){2-5} 
    \multirow{2}{*}{\textbf{Dialogue Gen.}}    & -     & -      & -     & -     \\
                                               & 8.4k  & 1.1k   & 11.9k & 16.5k \\
                                               & \tiny\textcolor{forestgreen}{+100\%} & \tiny\textcolor{forestgreen}{+100\%} & \tiny\textcolor{forestgreen}{+100\%} & \tiny\textcolor{forestgreen}{+100\%} \\
    % \midrule
    % \textbf{Books}                            & 27    & 28     & 15    & 15    \\
    % \textbf{Images}                           & 0.76k & 0.08k  & 0.93k & 1.78k \\
    \bottomrule
    \end{tabular}
\end{minipage}
\end{table}

\subsection{Qualitative details}
A notable example illustrating the design choices in the \textit{CoMix} annotations is found in the densely populated pages such as the "Naruto" page from PopManga (Figure \ref{fig:naruto-after}). This page features over 50 instances of the character Naruto, replicating himself using the ``Shadow Clone Technique''—a concept well-known among the fan base. However, in PopManga, such a page receives only partial annotations, primarily highlighting large-scale depictions and main characters. This selective approach underscores our focus on significant elements over exhaustive detailing, which aligns with our annotation strategy to emphasize clarity and relevance in highly complex scenes.

\begin{figure}[h]
\centering
\begin{subfigure}{.48\textwidth}
  \centering
  \includegraphics[width=0.9\linewidth]{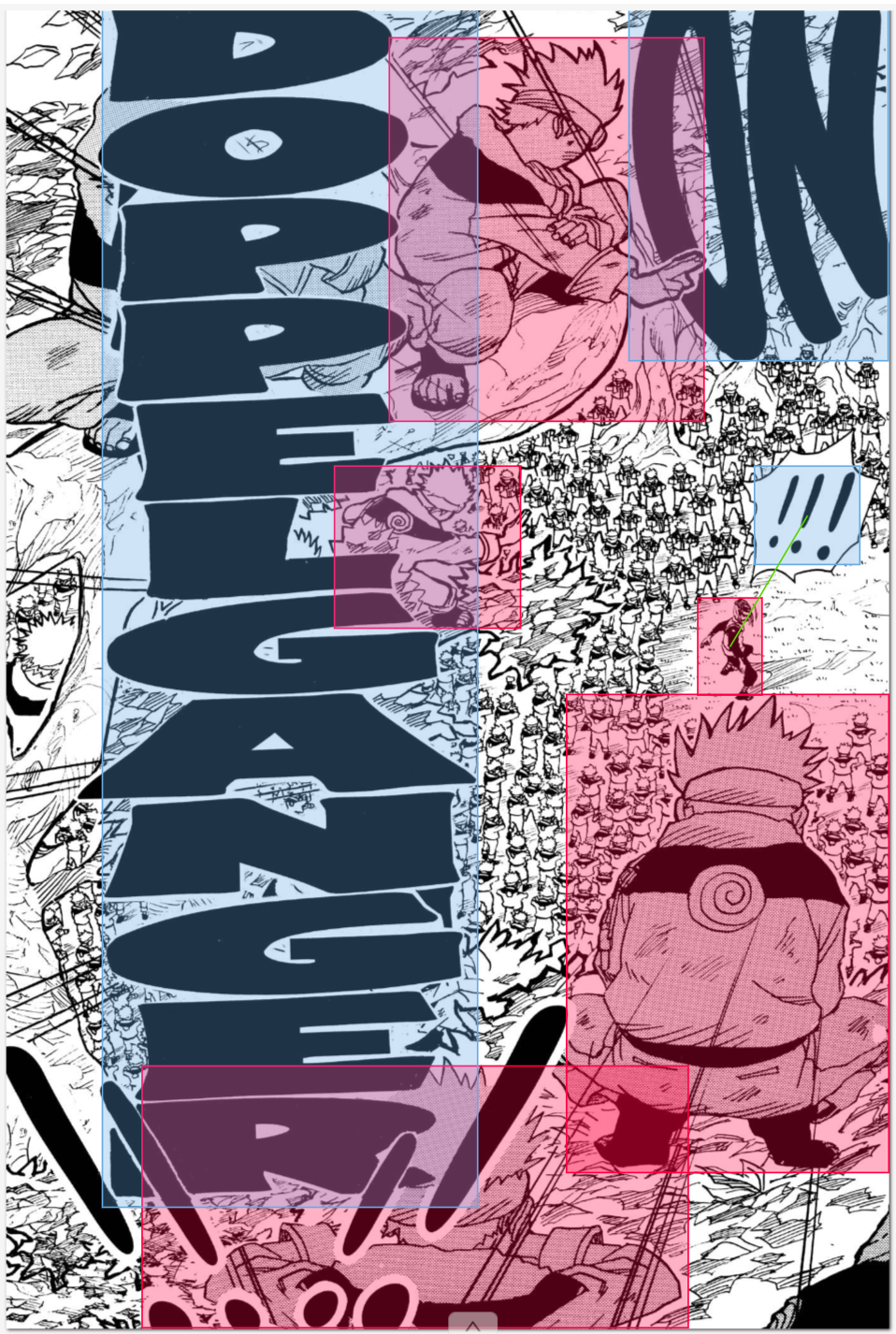}
\end{subfigure}%
\begin{subfigure}{.48\textwidth}
  \centering
  \includegraphics[width=0.9\linewidth]{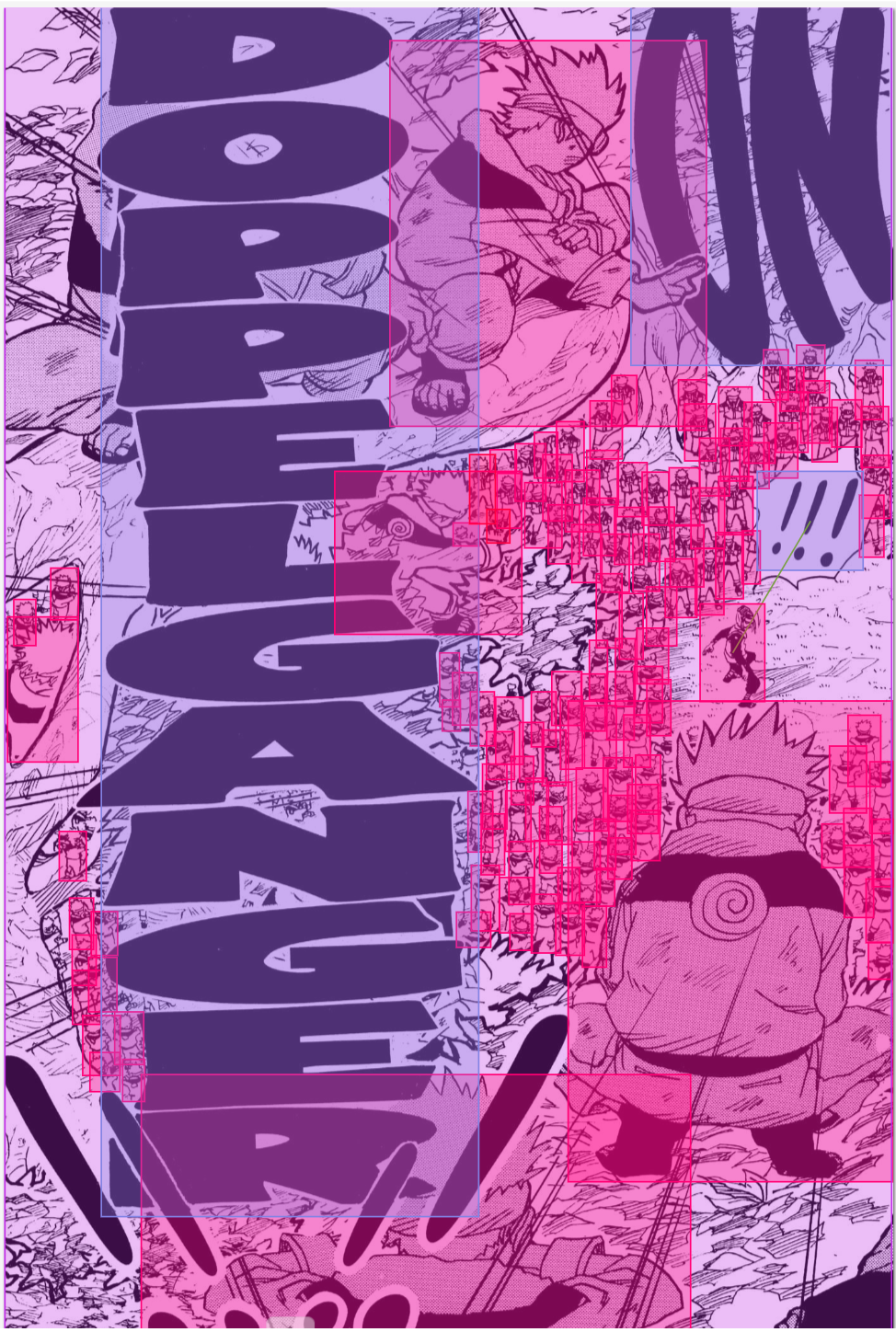}
\end{subfigure}
\caption{Image from ``Naruto'' (the ``Shadow Clone Technique'' from Chapter 2), annotated in the original PopManga dataset (left) and our \textit{CoMix} benchmark (right).}
\label{fig:naruto-after}
\end{figure}

\begin{figure}[h]
\centering
\begin{subfigure}{.48\textwidth}
  \centering
  \includegraphics[width=0.9\linewidth]{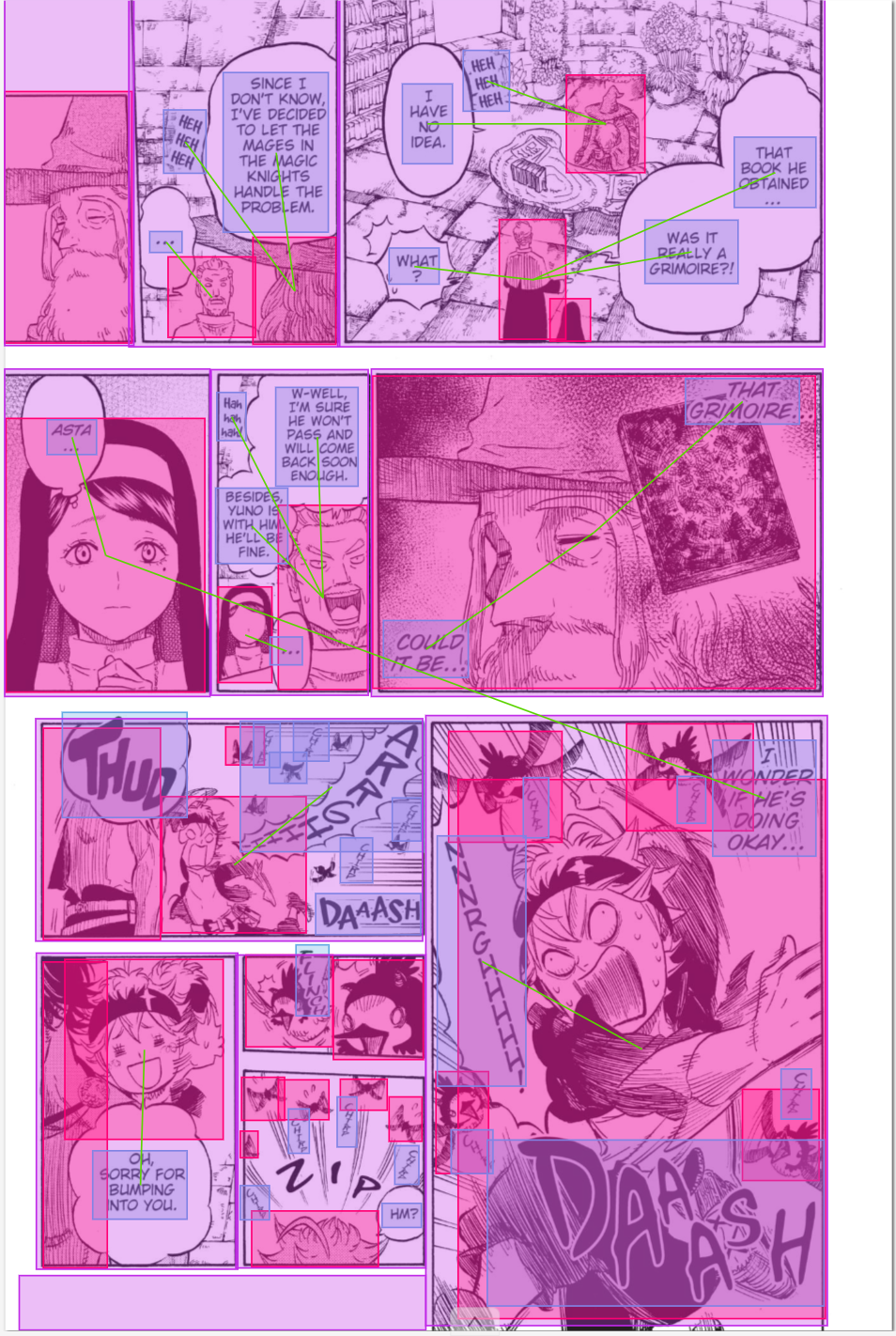}
\end{subfigure}%
\begin{subfigure}{.48\textwidth}
  \centering
  \includegraphics[width=0.9\linewidth]{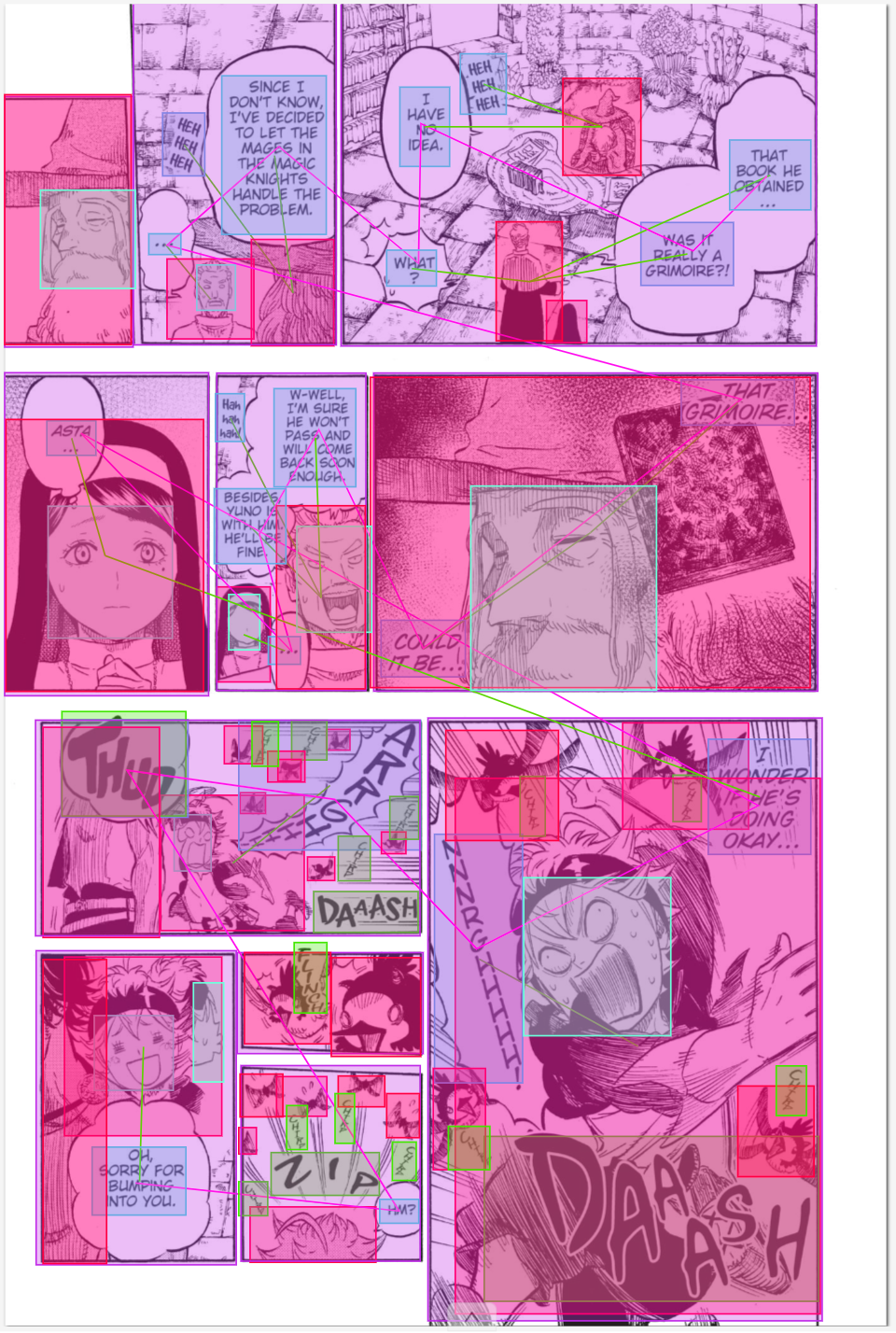}
\end{subfigure}
\caption{Image from ``Black Clover'' from the first chapters, annotated in the original PopManga dataset (left) and our \textit{CoMix} benchmark (right). In the \textit{CoMix} there are cleaned panel annotations, new faces boxes, corrected text to onomatopoeias annotations to and  but also }
\label{fig:naruto-after}
\end{figure}

\section{Data selection}
\label{sec:data-selection}
The \textit{CoMix} benchmark sees the presence of selected American golden-age comics from DCM \footnote{Digital Comics Museum at \href{https://digitalcomicmuseum.com}{https://digitalcomicmuseum.com}}. These documents have not been selected following the ``most downloaded'' principle, as instead is done in COMICS \cite{iyyer_amazing_2017}. This, as mentioned in the paper, is because the DCM website reports gross download numbers, not caring about single account downloads. This means, as reported in one comment on the website\footnote{Comment on the most downloaded book \href{https://digitalcomicmuseum.com/index.php?dlid=21158}{Wanted Comics 11 -JVJ}}, that repeatedly downloads by automatic bots count as single downloads, thus invalidating the global score validity. Instead, we want to guarantee the presence of various style comics, featuring as many characters as possible and whose characters are mostly present in these selected books. Thus we propose an algorithm to select the books with the required principles: ``Pow Selection Approach'', reported in the Algorithm \ref{algo:pow}.

\subsection{Books selection algorithm}
The book selection process emphasizes books containing characters frequently appearing across multiple books, suggesting greater narrative importance. This involves two key phases:

1. \textbf{Calculation of Shared-to-Unique Character Ratio:} For each book, a ratio is calculated based on the frequency of characters appearing within and in other books. To emphasize differences between books regarding character sharing, we squared the ratio.

2. \textbf{Selection of Top Books:} Books are ranked by their calculated ratios and the top 100 books with the highest scores are selected for further analysis.

The pseudocode is provided in Algorithm \ref{algo:1}. This algorithm ensures that the selected books reflect a broader narrative context, as it prioritizes those with characters that bridge multiple storylines. Such an approach is suitable for analyzing complex datasets where character interrelations are significant.

\begin{algorithm}
\caption{Book Selection Based on Character Sharing}
\label{algo:pow}
\begin{algorithmic}[1]
\Procedure{PowSelectionApproach}{ \textcolor{softviolet}{$book\_to\_characters$}, \textcolor{softblue}{pow}}
    \State {\textcolor{softviolet}{$book\_shared\_to\_unique\_ratio$}} $\gets$ \text{empty dictionary}
    \For{\textcolor{forestgreen}{$book\_id$}, \textcolor{forestgreen}{$characters$} in \textcolor{forestgreen}{$book\_to\_characters$}}
        \State $shared\_count \gets 0$
        \State $unique\_character\_count \gets \text{len( } characters$ \text{)}
        \For{\textcolor{forestgreen}{$other\_book\_chars$} in \textcolor{forestgreen}{$book\_to\_characters.values$}}
            \If{\textcolor{forestgreen}{$other\_book\_chars$} $\neq$ \textcolor{forestgreen}{$characters$}}
                \For{\textcolor{softviolet}{$char\_id$} in \textcolor{softviolet}{$characters$}}
                    \If{\textcolor{softviolet}{$char\_id$} in \textcolor{softviolet}{$other\_book\_chars$}}
                        \State $shared\_count \gets shared\_count + 1$
                    \EndIf
                \EndFor
            \EndIf
        \EndFor
        \State $ratio \gets \frac{shared\_count}{unique\_character\_count}$
        \State \textcolor{softviolet}{$book\_shared\_to\_unique\_ratio[book\_id]$} $\gets$ $ratio^{\textcolor{softblue}{pow}}$
    \EndFor
    \State \textbf{return} \textcolor{softviolet}{$book\_shared\_to\_unique\_ratio$}
\EndProcedure
\end{algorithmic}
\label{algo:1}
\end{algorithm}

\subsection{Comics books overview}
\begin{figure}[h]
\centering
\begin{subfigure}{.48\textwidth}
  \centering
  \includegraphics[width=0.9\linewidth]{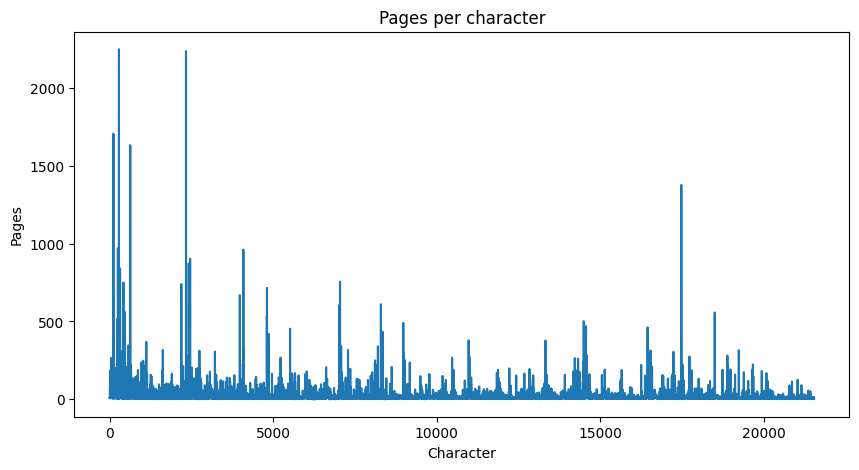}
\end{subfigure}%
\begin{subfigure}{.48\textwidth}
  \centering
  \includegraphics[width=0.9\linewidth]{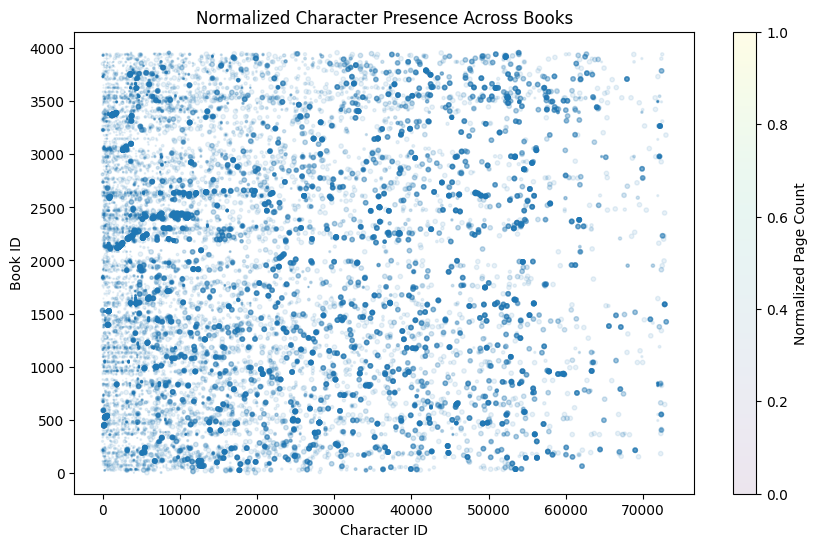}
\end{subfigure}
\caption{Overview of the characters per books. The bar-plot (left) represents the number of pages the characters appears in total. The right one, an overview of the presence of every character across books. Both graphs are generated using a first filtered subgroup of 4k books across the total 22k from DCM.}
\label{fig:100-out-22k}
\end{figure}

In this section, we provide details on the featured characters across the books and the selected ones. In Figure \ref{fig:100-out-22k}, on the left, a heatmap representing the occurrences of characters (y-axis) across different books (x-axis) considering the number of pages the character appears in. The number of pages is given by 

\begin{figure}[h]
\centering
\includegraphics[width=0.6\linewidth]{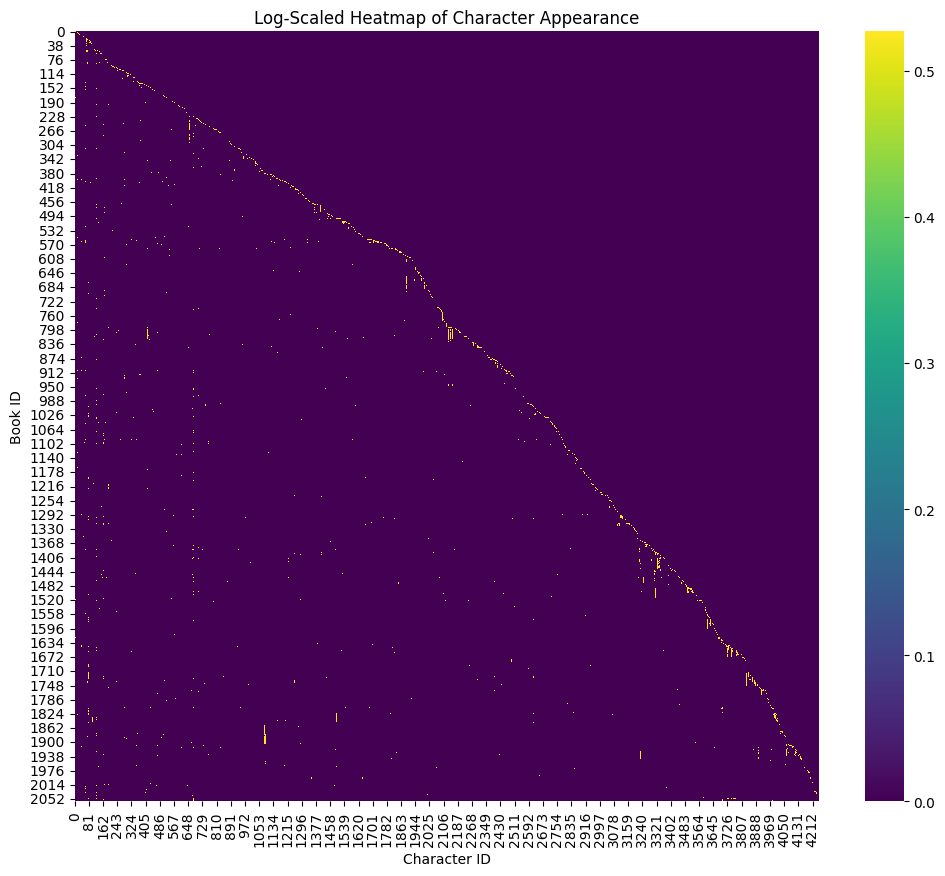}
\caption{Overview of the characters per book, after the filtering approach. As we can see from the vertical lines, characters appear in consecutive indexed books.}
\label{fig:heatmap}
\end{figure}

An additional step we have employed corresponds to filtering books that only contain one character and characters that are present in only one book (thus, the characters that have the spiking bar plot in Figure \ref{fig:100-out-22k}). We end up with a reasonable number of books (2k) for which more than 4k characters are present. \ref{fig:heatmap}. We calculated the Algorithm \ref{algo:1} on this collection of books.

\section{Detailed Results}
\label{sec:results}
% In this section, we provide extensive information about the baselines. 

\subsection{Detection}
In particular, for the detection baselines, we have fine-tuned two convolutional-based architectures, Faster R-CNN and YOLOv8, previously employed in Comics Object Detection, and utilize the available weights for DASS, a YOLOX-based model for character and face detection. We have also employed a transformer-based Magi model \cite{sachdeva_manga_2024} and a zero-shot transformer-based model for open-vocabulary detection called GroundingDino \cite{liu_grounding_2023}. The results, of the detection benchmarks, are reported in Table \ref{tab:results-det}. However, as some models are only trained for detecting a limited number of classes (DASS only detects faces and characters, while Magi does not detect faces). As a fair comparison, in Table \ref{tab:results-det-fair} we provide the detection metrics calculated only based on the detectable classes.

\begin{table}[ht]
\footnotesize
\begin{minipage}[b]{0.48\textwidth}
    \centering
    \caption{Results on \textbf{Detection task}, values reported are the mean Average Precision (mAP) over the four classes: Panel, Character, Text, Face.}
    \vspace{3mm}
    \label{tab:results-det}
    \begin{tabularx}{\linewidth}{rcccc|c}
    \toprule
    Models & \textbf{DCM} & \textbf{eBD} & \textbf{comics} & \textbf{Pop} & avg \\
    \midrule
    G.Dino$^{(4)}$ & 49,5 & 37,7 & 49,2 & 48,9 & 48,7 \\
    R-CNN$^{(4)}$ & {\ul 63,7} & 36,8 & \textbf{71,1} & 52,2 & \textbf{62,7} \\
    YOLO$^{(4)}$ & \textbf{65,2} & \textbf{55,9} & {\ul 64,9} & {\ul 54,7} & {\ul 61,3} \\
    DASS$^{(2)}$ & 41,3 & 21,8 & 34,2 & 20,9 & 30,4 \\
    Magi$^{(3)}$ & 63,2 & {\ul 42,4} & 57,2 & \textbf{62,1} & 58,9 \\
    \bottomrule
    \end{tabularx}
\end{minipage}\hfill
\begin{minipage}[b]{0.48\textwidth}
    \centering
    \caption{Results on \textbf{Detection task}, values reported are the mean Average Precision (mAP) over the predictable classes (DASS only detects Character and Face, and MAGI does not detect Face.)}
    \vspace{3mm}
    \label{tab:results-det-fair}
    \begin{tabularx}{\linewidth}{rcccc|c}
    \toprule
    Models & \textbf{DCM} & \textbf{eBD} & \textbf{comics} & \textbf{Pop} & avg \\
    \midrule
    G.Dino$^{(4)}$ & 49,5 & 37,7 & 49,2 & 48,9 & 48,7 \\
    R-CNN$^{(4)}$ & 63,7 & 36,8 & {\ul 71,1} & 52,2 & {\ul 62,7} \\
    YOLO$^{(4)}$ & 65,2 & {\ul 55,9} & 64,9 & {\ul 54,7} & 61,3 \\
    DASS$^{(2)}$ & {\ul 82,5} & 43,5 & 68,4 & 41,8 & 60,8 \\
    Magi$^{(3)}$ & \textbf{84,3} & \textbf{56,5} & \textbf{76,3} & \textbf{82,9} & \textbf{78,6}\\
    \bottomrule
    \end{tabularx}
\end{minipage}
\end{table}

\subsection{Speaker identification}
Regarding speaker identification, we provide two baseline results: using Magi or connecting the textbox with the closest character, within the panel. In Table \ref{tab:speaker} are reported the results for the two baselines, with metric $Recall@\text{\#text}$ as previously proposed \cite{li_manga109dialog_2023}. 

\begin{table}[h]
\centering
\caption{Results on \textbf{Speaker identification}, the task of connecting the textbox with the speaker character within the page. Values reported are the $R@\text{\#text}$ calculated on every page and averaged over the \textit{CoMix} datasets.}
\vspace{3mm}
\label{tab:speaker}
\begin{tabular}{rcccc|c}
\toprule
Models & \textbf{DCM} & \textbf{eBD} & \textbf{comics} & \textbf{Pop} & avg \\
\midrule
closest & \textbf{42,0} & \textbf{67,1} & \textbf{36,1} & 37,3 & \textbf{38,4} \\
Magi & 13,2 & 13,1 & 15,0 & \textbf{57,2} & 27,9 \\
\bottomrule
\end{tabular}
\end{table}

\subsection{Character Naming and Dialog generation}
For the tasks of character naming and dialog generation, as detailed in the paper, we introduce the Hybrid Dialog Score (\textit{HDS}), which combines the ANLS metric for assessing the accuracy of character names and the edit distance for evaluating the similarity between generated and ground truth dialog transcriptions. The methodology for computing the \textit{HDS} metric is elaborated in Algorithm \ref{algo:2}.

\begin{algorithm}
\caption{Hybrid Dialog Score}
\begin{algorithmic}[1]
\Procedure{EvaluateTranscription}{$model\_output, ground\_truth$}
    \State \textcolor{forestgreen}{$matches$} $\gets \text{find optimal matches}(model\_output, ground\_truth)$
    \State \textcolor{softviolet}{$tot\_ed$}, \textcolor{softblue}{$char\_name\_score$} $\gets 0, 0, 0$
    \For{each $(mo, gt)$ in \textcolor{forestgreen}{$matches$}}
        \State $edit\_dist \gets \text{calculate edit distance}(mo.text, gt.text)$
        \State \textcolor{softviolet}{$tot\_ed$} $\gets$ \textcolor{softviolet}{$tot\_ed$} $+$ $edit\_dist$  $/$ len($gt.text$)
        \State $anls\_score \gets \text{calculate ANLS}(mo.name, gt.name)$
        \State \textcolor{softblue}{$char\_name\_score$} $\gets$ \textcolor{softblue}{$char\_name\_score$} $+$ $anls\_score$
    \EndFor
    \State \textcolor{softviolet}{$tot\_ed$} $\gets 1 -$ \textcolor{softviolet}{$tot\_ed$}
    \State \textcolor{softblue}{$char\_name\_score$} $\gets$ \textcolor{softblue}{$char\_name\_score$} $/$ \text{len(\textcolor{forestgreen}{$matches$})}
    \State \Return \textcolor{softviolet}{$tot\_ed$}, \textcolor{softblue}{$char\_name\_score$}
\EndProcedure

\end{algorithmic}
\label{algo:2}
\end{algorithm}

Specifically, the results for Character Naming are summarized in Table \ref{tab:char-name}. Notably, the accuracy for the character naming task within the comic-style collection of the \textit{CoMix} dataset is particularly low for the Magi system. This is primarily because Magi does not explicitly solve the character naming task but rather assigns names such as "Char n", where $n$ is an incremental identifier for character clusters. In contrast, within the PopManga collection, Magi exhibits enhanced performance, even outperforming GPT-4. This improvement can be attributed to the simpler task of analyzing single-page manga-style comics, where dialog is less frequent and character interactions are less complex compared to traditional comics. Consequently, the predominance of unknown characters in manga-style comics allows Magi to achieve higher average ANLS scores than GPT-4.

\begin{table}[h]
    \centering
    \caption{
        \textbf{Results on Character Naming and Dialog Generation.} 
        The metrics reported are ANLS for Character Naming and minimum edit distance for Dialog Generation.
        Both metrics are calculated on every page and averaged over the \textit{CoMix} datasets.
    }
    \vspace{3mm}
    \begin{minipage}[b]{0.48\textwidth}
        \centering
        \small % Reduce font size
        \caption*{\textbf{Character Naming}. The ANLS metric measures the similarity of two strings, considering it wrong when more than 50\% do not agree.}
        \label{tab:char-name}
        \resizebox{\linewidth}{!}{ % Scale the table to fit the minipage width
            \begin{tabular}{rcccc|c}
                \toprule
                Models & \textbf{DCM} & \textbf{eBD} & \textbf{comics} & \textbf{Pop} & avg \\
                \midrule
                Magi & 9,0 & 7,0 & 8,0 & 45,0 & 19,76 \\
                GPT-4 & 54,0 & 37,0 & 58,0 & 28,0 & 47,11 \\
                \bottomrule
            \end{tabular}
        }
    \end{minipage}\hfill
    \begin{minipage}[b]{0.48\textwidth}
        \centering
        \small % Reduce font size
        \caption*{\textbf{Dialog Generation}. The minimum edit distance metric is calculated for every possible match between ground truth and predicted dialogs.}
        \label{tab:dialog}
        \resizebox{\linewidth}{!}{ % Scale the table to fit the minipage width
            \begin{tabular}{rcccc|c}
                \toprule
                Models & \textbf{DCM} & \textbf{eBD} & \textbf{comics} & \textbf{Pop} & avg \\
                \midrule
                Magi & 54,0 & 42,0 & 42,0 & 43,0 & 43,61 \\
                GPT-4 & 93,0 & 94,0 & 93,0 & 89,0 & 93,14 \\
                \bottomrule
            \end{tabular}
        }
    \end{minipage}
\end{table}

However, the dialog transcriptions generated by Magi are of inferior quality compared to those produced by GPT-4, which achieved an impressive average \textit{HDS} of 93.14\%. This disparity highlights the strengths and limitations of the respective systems in handling complex narrative elements within the \textit{CoMix} dataset.

\section{Additional Ethical Considerations}
\label{sec:add-ethical}
Alongside the copyright and consent information and the initial biases analysis featured in the main paper, we include information on automatic semantic harmful content analysis using cutting-edge Multimodal LLMs.

\subsection{Semantic Harmful Content}
The dataset has undergone a thorough inspection by the authors to ensure it is free from NSFW and offensive content. Throughout the inspection process, no content from the existing datasets (DCM, eBDtheque, PopManga) or the newly included American golden-age comics was deemed NSFW.

Regarding offensive content, it is acknowledged that comics from the 1950s, both American and European, occasionally used terms that were mildly offensive towards minorities and wartime adversaries (such as Japanese or German soldiers), which were culturally tolerated during that period. However, following the implementation of the Comics Code by the Comics Code Authority\footnote{\url{https://cbldf.org/comics-code-history-the-seal-of-approval}}, content depicting offensive themes or inspiring violence was strictly prohibited. Consequently, our selection is confined exclusively to comics published post 1954, ensuring all included works are compliant with the Comics Code. This careful curation supports the use of the dataset in diverse research and educational settings without risking exposure to inappropriate material.

Moreover, performing a full manual analysis of the semantic content and possible biases present in the \textit{CoMix} dataset is complicated, and prone to subjective biases of human evaluators. We have instead opted to perform an automatic analysis of the textual content at the panel level, using the Llama3-80B model. We provided the model with detailed descriptions of different semantic classes and classified each panel accordingly (see the classes in Table \ref{tab:classes}).

The model was provided with detailed descriptions of different semantic classes, as shown below:

\begin{tcolorbox}[
    colback=gray!5,
    colframe=gray!75,
    title=Model Prompt,
    fonttitle=\bfseries,
    boxrule=0.5pt,
    left=6pt,
    right=6pt,
    top=6pt,
    bottom=6pt
]
You are an image classifier trained to identify the offensiveness of comic balloon content. You will be provided with the extracted content of a panel, in a single row, which represents the concatenation of multiple balloons, following the reading order. Based on that, you must choose a number from -1 to 5 as one of the following:

<TABLE \ref{tab:classes}>

Examine the following text and determine the offensiveness score (-1, 0, 1, ..., 5). Respond only with the identified style, without any explanation.

<text>
\end{tcolorbox}

\begin{table}[h]
    \centering
    \caption{Classes provided to the model. The critical panels are the ones classified as 3, 4, and 5.}
    \vspace{3mm}
    \label{tab:classes}
    \begin{tabularx}{\linewidth}{c|c|X}
        \toprule
        \textbf{Code} & \textbf{Class} & \textbf{Short Description} \\
        \midrule
        -1 & Empty & The text is either empty or provides content which meaning is ‘empty text’. \\
        0 & Neutral & Completely neutral content with no offensiveness. \\
        & & \\
        1 & Informal & Casual content, maybe mild slang but not offensive. \\
        & & \\
        2 & Humorous/Sarcastic & Humorous content, unlikely to offend most people. \\
        & & \\
        3 & Sexual/Angry/Violent & Sexual, angry, or violent content, potentially inappropriate for some but not offensive. \\
        4 & Offensive & Insulting content, but not racist or sexist. Just clearly offensive. \\
        5 & Highly Offensive/Racist/Sexist & Extremely racist, sexist, or offensive. \\
        \bottomrule
    \end{tabularx}
\end{table}

From the initial analysis, we were able to obtain the results shown in Table \ref{tab:results}.

\begin{table}[]
    \centering
    \caption{Semantic harmful content analysis.}
    \vspace{3mm}
    \label{tab:results}
    \begin{tabular}{c|l|l}
    \toprule
         \textbf{code} & \textbf{class} & \textbf{percentage \%}  \\
    \midrule
    -1 & Empty & 23.9 \% \\
    0 & Neutral & 13.3 \% \\
    1 & Informal & 14.5 \% \\
    2 & Humorous/Sarcastic & 34.2 (34.7) \% \\
    3 & Sexual/Angry/Violent & 12.7 (13.1) \% \\
    4 & Offensive & 1.6 ( 1.2 ) \% \\
    5 & Highly Offensive/Racist/Sexist & 0.8 (0.3) \% \\
    \bottomrule
    \end{tabular}
\end{table}

\begin{figure}
    \centering
    \includegraphics[width=0.5\linewidth]{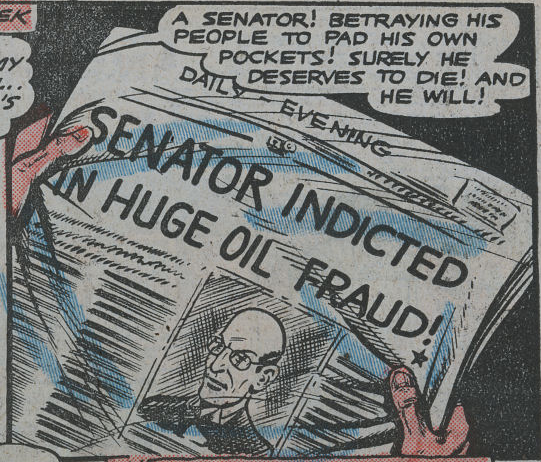}
    \caption{Example of a “Highly Offensive/Racist/Sexist” image (based on textual information) that, instead, belongs to “violent” or “incitement to violence” class.}
    \label{fig:example}
\end{figure}

We subsequently manually checked 370 panels to verify these results (from classes 3,4,5). We found that many of these are actually less (or not) offensive than the model classifications. An example of a panel being classified as “Highly Offensive/Racist/Sexist” is provided in Figure \ref{fig:example}. 

We updated the percentage in Table \ref{tab:results} considering these manual verifications. These statistics are meant to be indicative of the semantic content of our dataset, while we consider that a full, detailed study is out of the context of this manuscript.

\end{document}